\documentclass[journal,twoside,web]{ieeecolor}
\usepackage{generic}
\usepackage{cite}
\usepackage{amsmath,amssymb,amsfonts}
\usepackage{algorithmic}
\usepackage{graphicx}
\usepackage{algorithm}
\usepackage{hyperref}
\usepackage{rotating}
\usepackage{textcomp}
\usepackage{color,soul}
\usepackage{multirow}
\usepackage{booktabs}
\usepackage{pifont}
\usepackage{amsmath}
\usepackage{array} 
\usepackage{caption}
\usepackage{subcaption}

\let\labelindent\relax

\usepackage{enumitem,kantlipsum}

\def\BibTeX{{\rm B\kern-.05em{\sc i\kern-.025em b}\kern-.08em
    T\kern-.1667em\lower.7ex\hbox{E}\kern-.125emX}}
\begin{document}

\title{How Deep is your Guess? A Fresh Perspective on Deep Learning for Medical Time-Series Imputation}

\author{Linglong Qian, Hugh Logan Ellis, Tao Wang, Jun Wang,  Robin Mitra, Richard Dobson and Zina Ibrahim$^\dagger$
\thanks{$^\dagger$Correspondence to Zina Ibrahim $<$zina.ibrahim@kcl.ac.uk$>$}
\thanks{Linglong Qian, Tao Wang and Hugh Logan Ellis are with the Department of Biostatistics and Health Informatics, King's College London, SE5 8AF London, UK}
\thanks{Jun Wang is with the Department of Computer Science, University of Warwick, CV4 7AL Coventry, UK and also with the Department of Informatics, King’s College London, SE5 8AF, London, UK}
\thanks{Robin Mitra is with the Department of Statistics, University College London, WC1E 7HB, London, UK}
\thanks{Zina Ibrahim and Richard Dobson are with the Department of Biostatistics and Health Informatics, King's College London, SE5 8AF, London UK and also with the Institute of Health Inforamtics, University College London, WC1E 6BT London, UK}
\thanks{Linglong Qian, Tao Wang, Zina Ibrahim and Richard Dobson are also with the King's Institute for AI}
\thanks{Linglong Qian and Zina Ibrahim are also with PyPOTS Research}
}

\maketitle

\begin{abstract}

We present a comprehensive analysis of deep learning approaches for Electronic Health Record (EHR) time-series imputation, examining how architectural and framework biases combine to influence model performance.
Our investigation reveals varying capabilities of deep imputers in capturing complex spatiotemporal dependencies within EHRs, and that model effectiveness depends on how its combined biases align with medical time-series characteristics. Our experimental evaluation challenges common assumptions about model complexity, demonstrating that larger models do not necessarily improve performance. Rather, carefully designed architectures can better capture the complex patterns inherent in clinical data. The study highlights the need for imputation approaches that prioritise clinically meaningful data reconstruction over statistical accuracy.
Our experiments show imputation performance variations of up to 20\% based on preprocessing and implementation choices, emphasising the need for standardised benchmarking methodologies. Finally, we identify critical gaps between current deep imputation methods and medical requirements, highlighting the importance of integrating clinical insights to achieve more reliable imputation approaches for healthcare applications.

\end{abstract}


\section*{Nomenclature}

{\small

\begin{tabular}{@{}p{2cm}@{}p{\dimexpr\textwidth-2.5cm\relax}}
BRITS   & Bidirectional Recurrent Imputation  \\
   &  for Time Series \\
CNN     & Convolutional Neural Network \\
CSBI    & Conditional Score-Based Imputation \\
CSDE    & Conditional Stochastic Differential Equation \\
CSDI    & Conditional Score-based Diffusion Imputation \\
CRU     & Continuous Recurrent Unit \\
\end{tabular}
}

{\small
\begin{tabular}{@{}p{2cm}@{}p{\dimexpr\textwidth-2.5cm\relax}}
GAN     & Generative Adversarial Network \\
GLIMA   & Global and Local Imputation with  \\
GLIMA   & Multi-directional Attention \\

GNN     & Graph Neural Network \\
GP      & Gaussian Process \\
GRUD    & Gated Recurrent Unit with Decay \\
GRU     & Gated Recurrent Unit \\
HI-VAE  & Heterogeneous Incomplete  \\
  & Variational Autoencoder \\
ICU     & Intensive Care Unit \\
LSTM    & Long Short-Term Memory \\
MAE     & Mean Absolute Error \\
MAR     & Missing at Random \\
MCAR    & Missing Completely at Random \\
MDN     & Mixture Density Network \\
MIWAE   & Multiple Imputation with \\
   & Variational Autoencoders \\
MLP     & Multi-Layer Perceptron \\
MNAR    & Missing Not at Random \\
MRNN    & Multivariate Recurrent Neural Network \\
MSE     & Mean Squared Error \\
MTSIT   & Multivariate Time Series Imputation \\
  & with Transformers \\
NAM     & Normalisation After Masking \\
NBM     & Normalisation Before Masking \\
ODE     & Ordinary Differential Equation \\
PyPOTS  & Python Partially Observed Time Series \\
RNN     & Recurrent Neural Network \\
SAITS   & Self-Attention-based Imputation for Time Series \\
SDE     & Stochastic Differential Equation \\
SSSD    & State Space Score-based Diffusion \\
VAE     & Variational Autoencoder \\
V-RIN   & Variational Recurrent Imputation Network \\
YAIB    & Yet Another ICU Benchmark
\end{tabular}
}



\section{Introduction}
\label{sec:introduction}

\IEEEPARstart{T}{he} ever-increasing rate of data generation in electronic health records (EHRs) has enabled sophisticated deep learning architectures performing patient-centered predictive tasks \cite{mllandscape,Yang2022DeepMPMAM,Si2020DeepRL,islam,deeppredictionreview2}. Those range from mortality prediction \cite{mortality} to early detection of adverse events \cite{cardiacarrest} and disease progression modeling. The impact of these models is evident in different settings. In intensive care, early warning systems processing vital signs and laboratory values can predict deterioration before clinical manifestation. Here, deep learning models have demonstrated superior accuracy in predicting critical events such as cardiac arrest compared to traditional scoring systems \cite{lee2018deep}. In sepsis detection, where each hour of delayed treatment increases mortality by 7-8\%, deep learning models have been associated with a 5.0\% increase in sepsis bundle compliance and a 1.9\% decrease in hospital mortality \cite{deepsepsis}.  In cardiology, algorithms analysing ECG measurements and vital signs have been shown to detect subtle patterns indicating impending arrhythmias \cite{che2018recurrent}, enabling preventive interventions.
	
Regardless of the predictive task, missing data is a major issue \cite{garcia2010pattern}, making imputation a necessary step before machine learning algorithms can be applied. However,  EHR data is naturally complex, with many of the variables included clinical time-series being predominantly missing not at random  \cite{luo2022evaluating, mnarnature} and established correlation between EHR missingness patterns and clinical workflow and practices \cite{correlation1, correlation2, correlation3}. Given that imputation quality has been shown to significantly impact the performance of downstream clinical predictive models \cite{che2018recurrent}, capturing the underlying data properties during imputation is essential to preserve the clinical meaning of the data and to maintain the reliability and clinical applicability of the subsequent downstream task \cite{mitra2023learning}.


Recent advances in deep learning techniques have outperformed traditional statistical and machine learning imputation methods when applied to large and heterogeneous multivariate time-series datasets \cite{wang2024deep}. These models, herein referred to as \emph{deep imputers}, learn complex non-linear patterns and high-dimensional relationships with varying degrees of distributional flexibility and without requiring explicit feature engineering. Such flexibility is particularly valuable in dealing with EHR datasets, where intricate feature dependencies render feature engineering difficult and the underlying distribution may be unknown or highly variable \cite{complex}. 

Recent literature has examined deep learning approaches for time series imputation in healthcare through two key reviews. Liu et al. \cite{reviewhealthcare} survey reported performance across imputation methods, comparing deep imputers to traditional statistical and machine learning models. In contrast, Kazijevs \& Samad \cite{reviewhealthcare2} conduct a systematic experimental evaluation of eight deep imputers across different missingness scenarios and data conditions. While Kazijevs \& Samad provide valuable benchmarking insights, neither review examines the theoretical foundations that influence model performance or connects these to the complex characteristics of EHR data. This gap is particularly significant given the substantial variations in reported performance across different tasks, architectures, implementation approaches and training data \cite{tsibench}. Furthermore, deep imputers vary considerably in their data preprocessing steps and pipeline implementations, which can significantly affect imputation performance. In this regard, our review revealed a variety of data processing and masking techniques used to simulate missingness during experimental evaluation; these have not yet been systematically studied, leaving a critical gap in our understanding of the factors that lead to the performance metrics reported in reviews. More importantly, the current level of scrutiny of the various types of deep imputers is insufficient to make an informed choice of the appropriateness of a given model for a specific dataset.
 

We have therefore endeavoured to ask the following questions to support thoughtful assessment of the suitability of a given deep imputer for a given task: \textbf{a)} How well are the complex dependencies of the dataset (temporal and cross-sectional correlations) reflected in the design of a given deep imputer? \textbf{b) }How effectively do current deep imputers account for the distinct sampling patterns in EHR datasets, particularly that those patterns are governed by clinical workflows and patient characteristics? \textbf{c) }How do specific implementation choices (masking strategies, normalisation setup, initialisation methods, and pipeline design) impact model evaluation rigor and performance reliability? \textbf{d)} how consistent are the current practices used to benchmark deep imputers' performance? and \textbf{e) }What are the open questions that need to be addressed to ensure imputation approaches align with the practical requirements of clinical applications?

Our review provides a theoretical grounding for the observed performance of different deep imputers reported in the literature and experimentally evaluates practical implementation decisions that can drastically influence imputation performance. Our goal is to enhance the understanding of design choices, inform model and architecture selection, and refine evaluation methodologies, calling for controlled experimental environments and open benchmarking environments. We build on existing reviews by analysing deep EHR imputers along three dimensions: 

\begin{enumerate} [wide, label=\textbf{\arabic*}), labelwidth=!]
    \item \textbf{Taxonomy: }We present a theoretically-backed organisation of existing methods based on their inductive bias \cite{inductivebias} - the inherent assumptions these models make about how data is represented and processed, how imputations are generated and how a model captures complex dependencies and missingness patterns. Our taxonomy is hierarchical, examining how basic architectural and framework biases interact with design choices to give rise to higher-level characteristics of a given deep imputer model and distinct approaches for handling the complexity of medical time series.
    
    \item \textbf{Targeted Performance Comparison:} Through experimental evaluation of state-of-the-art deep imputers in an open and controlled environment, we study the relationship between model complexity and imputation accuracy. We also show that the effectiveness of an imputation model depends on the alignment between its combined biases and EHR data characteristics. Finally, we analyse model sensitivity to missingness mechanisms and different flavours of evaluation settings. Our findings a) challenge the assumption that larger models lead to better performance, b) establish salient relationships between design choices and reported performance, and c) highlight how common, yet often inadequate, evaluation practices can obscure benchmarking results. 

    \item \textbf{Gaps in EHR Imputation Research:} We identify critical gaps between current paradigm of deep imputers and the requirements imposed by the domain, particularly in uncertainty quantification, the incorporation of domain knowledge and our current understanding of EHR missingness. These findings highlight the importance of aligning the field's research directions with medical insights to develop more reliable imputation strategies.
\end{enumerate}

Our review aims to provide practical insights for researchers working with medical time series, helping them select appropriate models by considering computational costs and model sensitivity to data conditions. We also offer guidance for developers of novel imputation models that can operate under the complex settings EHRs, while highlighting important directions for future research in clinical data imputation.

Our review is structured as follows: We first provide necessary background by \textbf{a)} reviewing how EHR data collection leads to unique characteristics and missingness properties, and \textbf{b)} offering an overview of inductive bias in deep learning, highlighting major biases in known architectures and neural frameworks. We then detail our taxonomy, outlining how basic architectural and framework biases manifest into distinct characteristics of deep imputers, and explaining how these translate into advantages and limitations with respect to EHR data characteristics.
Our experimental evaluation follows, benchmarking eight deep imputers using PyPOTS, an open-source toolkit that provides standardised tools for multivariate time series imputation and model benchmarking. Our experimental results highlight important gaps in existing evaluation methodologies and provide insights into model selection and the tradeoffs between model complexity and imputation performance. Finally, we identify open research questions of particular importance for deep EHR imputation research.

\section{Background}

\subsection{Characteristics of EHR Data} \label{sec:ehrdata}

\begin{table*}[!t]
\renewcommand{\arraystretch}{1.3}
\caption{Characteristics and dependencies of clinical time series and the resulting missingness mechanisms. \textbf{MCAR:} missing completely at random; \textbf{MAR:} missing at random; \textbf{MNAR:} missing not at random.}
\label{tab:background}
\resizebox{\linewidth}{!}{
\begin{tabular}{lllllll}
\midrule
\textbf{Modality} & \textbf{Clinical} & \textbf{Clinical} & \textbf{Recording} & \textbf{Temporal} & \textbf{Cross-sectional} & \textbf{Missingness} \\
\textbf{} & \textbf{Category} & \textbf{Examples} & \textbf{Patterns} & \textbf{Dependencies} & \textbf{Correlations} & \textbf{Patterns} \\
\midrule
\multirow{2}{*}{\textbf{Continuous}} & 
Waveform & Invasive vital & Millisecond & Short-term: & Correlation with & MNAR: equipment \\
& Signals & sign monitoring, &  & beat-to-beat variations & derived vitals & disconnection, e.g. \\
&  & ECG signals &  &&  &  patient bathing\\

\midrule
\multirow{2}{*}{\textbf{High-Frequency}} &
Derived & Heart rate, & Every 1-5 minutes; & Short: acute & Strong correlation & MAR: during \\
& Vitals & Blood Pressure,& setting-dependent & physiological& within vital panels & in-hospital patient  \\
&  & Temperature &  &  responses &  & transport \\

\midrule
\multirow{10}{*}{\textbf{Discrete}} &
Laboratory & Full blood count, & Treatment-dependent; & Short: acute & Strong correlation & MAR: depends on \\
& Tests & Kidney function & ordered in panels & organ dysfunction & within panels & disease severity \\
\cline{2-7}
& Ventilator & FiO2, PEEP, & Protocol-driven & Short: ventilation & Strong correlation & MCAR/MNAR: \\
& Data & Tidal volume & sampling & adjustments & within parameters & technical/clinical \\
\cline{2-7}
& Medications & Antibiotics, & Event-based, & Short: treatment & Correlation with & Structured: \\
& & Vasopressors & effect cycles & response & lab results & protocol-based \\
\cline{2-7}
& Procedures & Surgery, & Irregular, & Short: acute & Strong correlation & Structured: \\
& & Endoscopy & protocol-driven & intervention effects & with vitals & protocol-based \\
\cline{2-7}
& Clinical & Assessments, & Daily/shift-based & Long: disease & Correlation with & MAR: severity \\
& Notes & Care plans & updates & progression & multiple measures & dependent \\
\midrule
\multirow{6}{*}{\textbf{Ordinal}} &
Clinical & PHQ9 & Daily/weekly & Short: acute & Correlation with & MAR: status \\
& Scores &  & or shift-dependent & deterioration & vital signs & dependent \\
\cline{2-7}
& Disease & Cancer stage, & Monthly/quarterly & Long: disease & Correlation with & Structured: \\
& Staging & Heart failure class & updates & progression & multiple markers & protocol-based \\
\cline{2-7}
& Risk & Mortality, & Admission/ & Short/Long: risk & Complex correlation & Structured: \\
& Assessments & Readmission risk & discharge timing & evolution & with multiple vars & event-based \\
\bottomrule
\end{tabular}}
\end{table*}


The primary function of EHRs is to support clinical care. EHR data accumulates during routine healthcare activities, forming patient trajectories that clinicians rely on for decision-making. These trajectories cover various dimensions of a patient's health status, such as demographics, diagnoses, medications, procedures, vital signs, test results, imagery and clinical text notes in a sequence of visits. Using the time series generated by this data resource to inform predictive tasks requires dealing with a unique set of characteristics; those are summarised in Table \ref{tab:background}.

The multidimensional coverage of EHR time series renders them inherently \textbf{multimodal}, capturing patient health dynamics through \textit{continuous measurements} (e.g. ECG signals), high- and low-frequency \textit{discrete events} (e.g. heart rate, medication changes) and \textit{ordinal data} (e.g. cancer stages). These multimodal variables are recorded \textbf{asynchronously} at intervals determined by device capabilities, clinical events, and protocols, among others \cite{jensen2012mining}. Device resolution dictates high-frequency measurements, such as automated recordings by blood pressure cuffs every five minutes in intensive care units, compared to capillary blood glucose measurements via a glucometer 4-6 times daily \cite{moss2017continuous}. Event-based recordings are primarily driven by treatment requirements, exemplified by insulin administration before meals with subsequent adjustments based on glucose readings \cite{glucose}. Drug recording patterns are particularly influenced by their effect cycles—the temporal sequence from administration through onset, peak effect, duration, and decline—creating predictable physiological timelines \cite{moskovitch2009medical}. Laboratory tests, such as daily complete blood counts during acute leukaemia treatment or troponin tests if a patient experiences chest pain, are treatment-dependent and are performed based on anticipated clinical needs \cite{pivovarov2014identifying}. Additionally, institutional protocols impose structured recording intervals, such as nursing assessments clustering around 12-hour shift end intervals, physician notes predominantly recorded during morning rounds, and risk assessments performed at admission and discharge. Some measurements follow strict periodic schedules, such as quarterly cancer staging during chemotherapy, whilst others occur irregularly in response to acute clinical needs, as seen with emergency surgical procedures.

Because many health parameters are inherently connected, clinical variables exhibit complex interdependencies and their missingness patterns often carry meaningful information about patient state and care processes. Many variables are \textbf{cross-sectionally correlated}, which can introduce redundancy by overemphasising certain features. For example, clinical practice often dictates ordering test panels and laboratory tests are seldom requested in isolation; ordering electrolyte tests typically includes kidney function tests, and calcium tests require concurrent albumin measurements to adjust calcium levels \cite{pivovarov2014identifying}.  EHR time-series correlations also extend across the temporal dimension, encompassing \textbf{short-term and long-term temporal dependencies} that hold significant meaning within a patient's trajectory. Immediate physiological responses reveal acute body reactions, while prolonged interventions or chronic conditions manifest long-term effects. For example, a heart attack typically results in immediate-term changes in vital signs (including blood pressure and heart rate), intermediate-term changes in biomarkers (such as troponin and renal function markers), and if it progresses to heart failure, long-term alterations in multiple physiological parameters \cite{heartattack}.  EHR correlations introduce an additional level of complexity for clinical predictive models, as clinical insights often lie in the extreme values of the complex data distributions of clinical variables, such as abnormally high or low blood sugar. An imputation model, therefore, needs to accurately discern those informative outliers which mark physiological events, from noise.

\subsection{Inductive Bias Across Neural Architectures \& Frameworks}

Our view of deep imputers is grounded on the notion of \textbf{inductive bias} \cite{inductivebias}, which refers to the set of preferences, priors or assumptions a deep learning model inherently makes, guiding the learning process to reduce the space of possible solutions by prioritising one solution over another, independent of the observed data. Therefore, before we examine EHR imputation in the literature, we equip the reader with the types of inductive bias known to existing architectures and frameworks that have been used in EHR deep imputers.


\begin{figure}[!htbp]
\centering
\includegraphics[width=\linewidth, height=7.2in]{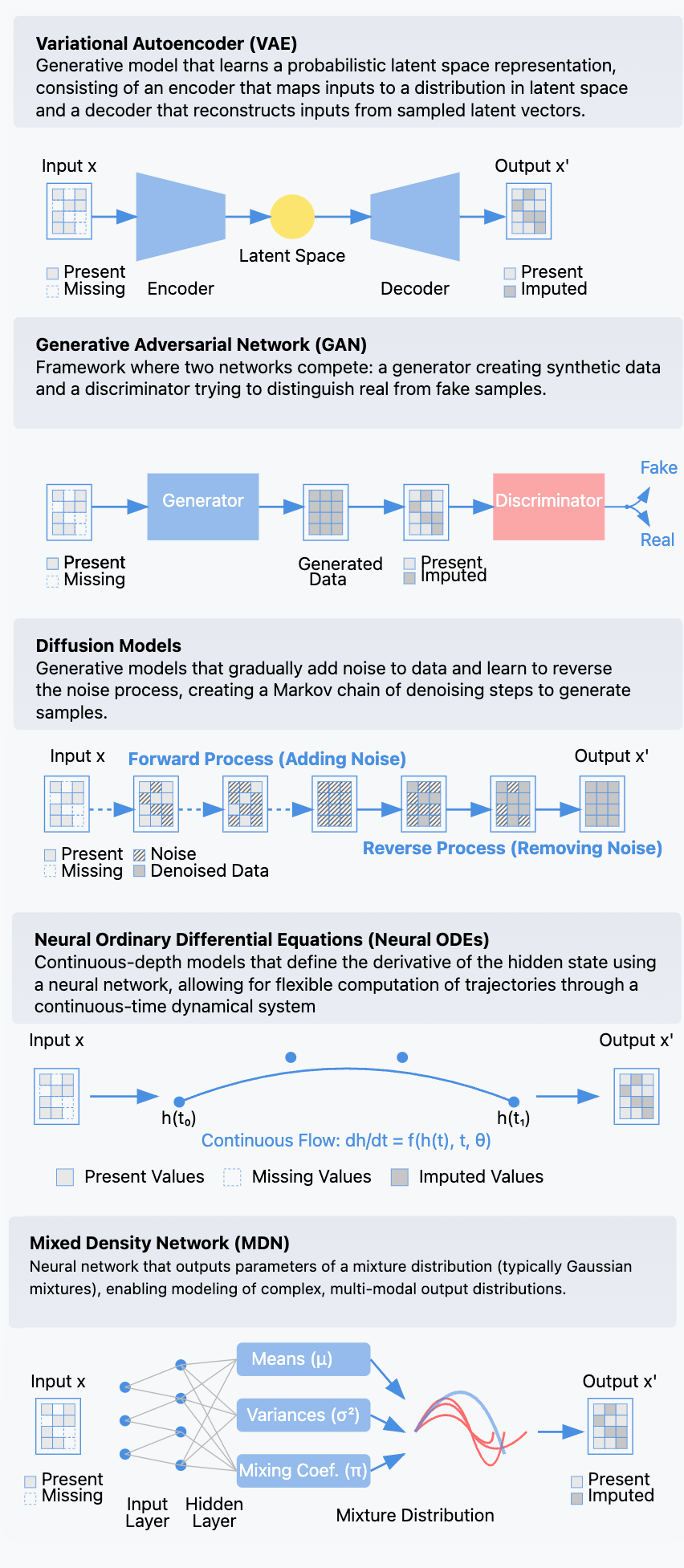}
\caption{A conceptual overview of generative frameworks used in medical time-series imputation. }
\label{fig:frameworks}
\end{figure}

Neural architectures differ in the way they \emph{represent and process data}. \textbf{Recurrent Neural Networks (RNNs)} are intuitively suited to handle temporal sequences\cite{medsker2001recurrent}. Their inherent inductive bias favours learning short-term temporal dependencies between variables over time by maintaining internal (hidden) states across time steps in their recurrent architectures. \textbf{Transformers} deviate from traditional sequence processing methods by capturing long-range dependencies through self-attention, with an inherent bias towards global contextuality \cite{vaswani2017attention}.  \textbf{Convolutional Neural Networks (CNNs)} possess an inductive bias towards cross-sectional patterns, useful for detecting acute physiological changes that can mark important clinical events, e.g. detecting signs of leukemia from white blood cell morphology. Finally, \textbf{Graph Neural Networks (GNNs)} \cite{zhou2020graph} exhibit a relational graph bias, representing complex interdependencies through message-passing algorithms on node-edge topologies, enabling them to model the complex relational structures between interconnected health indicators.

Modern deep imputers employ a variety of generative learning frameworks to improve their imputation; those are summarised in Figure \ref{fig:frameworks}. Generative frameworks diverge in their approach to data generation and uncertainty representation. \textbf{Variational Autoencoders} (VAEs) assume that data is generated from a latent space, and aim to learn a distribution, usually Gaussian \cite{mackay1998introduction}, capturing the underlying structure of the data. As such, VAEs are inherently probabilistic models biased by the assumed distribution. \textbf{Mixture Density Networks (MDNs)} overcome the single-distribution bias by assuming that the data is generated from a mixture of probability distributions. They directly capture imputation uncertainty through a mixture of weights and variances in the assumed distributions. 

\textbf{Generative Adversarial Networks (GANs)} \cite{creswell2018generative} take distribution complexity further by being inherently biased towards generating realistically diverse data distributions. This is done through an adversarial setup where the generator aims to fool the discriminator. While this enables effective generation of multivariate time-series, GANs often struggle with mode collapse, producing a limited variety of common patterns while missing rare but clinically significant events. Furthermore, GANs inherently lack direct mechanisms to quantify uncertainty within the imputations, and their application to establishing confidence in the generated data is still in its early stages \cite{oberdiek2022uqgan}. \textbf{Neural Ordinary Differential Equations (Neural ODEs)} \cite{chen2018neural} describe the temporal evolution of the data using differential equations, using the learned function to simulate the dynamics of the data over time. ODE models favour continuous data transitions aligning with the functions learned by the model \cite{schirmer2022modeling,park2021neural}. While their continuity bias can help handle irregular sampling in EHR data \cite{rubanova2019latent}, it also prevents them from capturing sudden physiological changes and discrete interventions common in healthcare settings (e.g., sudden onset of symptoms or treatment effects). Although uncertainty can be incorporated as a stochastic process, solving these equations is computationally intensive and highly sensitive to initial conditions. 
Finally, \textbf{Diffusion Models} \cite{ho2020denoising, song2020score} employ a stochastic process to gradually generate synthetic data, progressing from random noise towards distributions that mimic the observed data. Diffusion models assume that data evolves over time according to a gradual process. However, similar to Neural ODEs, the gradual denoising bias may cause diffusion models to miss or smooth over clinically significant sudden changes and may struggle to capture complex dependencies between clinical variables that change at different scales. Although diffusion models do not provide an explicit mechanism to quantify uncertainty, they can estimate uncertainty by measuring the divergence between predicted and observed data distributions at each time step.

\begin{figure*}

\centering
\includegraphics[width=14cm]{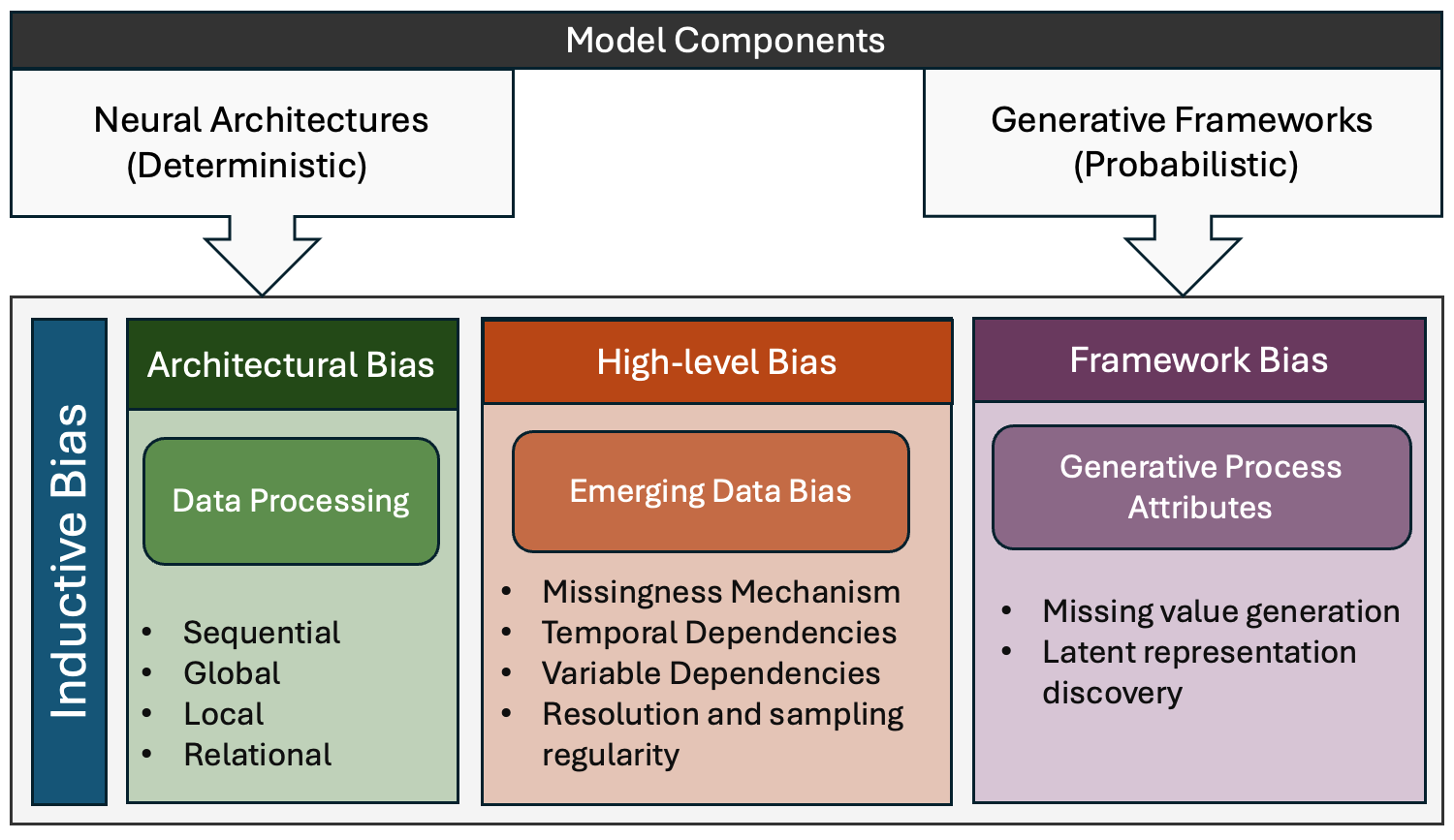}
\caption{The hierarchy of dimensions of inductive bias governing the behaviour of deep imputation models} \label{fig:dimensions}
\end{figure*}

\section{Taxonomy of Deep Imputers}

Our multidimensional exploration of the literature organises deep imputers by examining their approaches to handling the complex characteristics of EHR data discussed in Section \ref{sec:ehrdata}. Backed by the knowledge that inductive bias significantly influences a model's behaviour and generalisation capabilities  \cite{inductivebias}, our taxonomy is formulated using the following observations, which are illustrated in Figure \ref{fig:dimensions}.


\begin{enumerate} [wide, label=\textbf{\arabic*})]

   \item \textbf{Component Integration:  }Many modern deep imputers combine neural network \textbf{architectures} with probabilistic generative \textbf{frameworks}, each serving a complementary role with distinct inductive biases. Architectures encode structural assumptions about data patterns and determined how data is processed, while frameworks determine preferences about how missing values are modeled and how the underlying distribution of the data is discovered, guiding the imputation process towards plausible generalisations that capture and replicate complex data distributions. The resulting synergy of preferences shapes a deep imputer's attempt to generate data which reflects realistic and clinically relevant patterns. 
   \item    \textbf{Hierarchical Organisation:} An imputer's inductive biases accumulate: design modifications influence how fundamental architectural and framework biases inform higher-level preferences about data dependencies and patterns. This hierarchy guides systematic model development by connecting design choices to underlying assumptions about the data's complexity and missingness patterns.
   \item \textbf{Model-Bias Alignment}: We hypothesise that a deep imputer's performance depends on how well its fundamental and higher-level biases align with dataset characteristics. Mismatches between architectural, framework, or high-level biases and data properties can limit model effectiveness, guiding systematic choices in model design.

\end{enumerate}
 
Our organisation of the literature is shown in Tables \ref{tab:overview} and \ref{tab:higher}. Our selection criteria for analysis and benchmarking are as follows: 

\begin{enumerate}  [wide, label=\textbf{\arabic*}), labelwidth=!]
\item \textbf{Survey and Analysis:} deep learning methods designed for multivariate time series, evaluated on at least one EHR dataset. Excluded statistical and traditional machine learning methods and those without EHR benchmark validation. 
\item \textbf{Experimental Evaluation:} To ensure standardised evaluation and full control over experimental conditions, we perform our experiments using the \texttt{PyPOTS} library (more details in the experimental evaluation section). We therefore only evaluate deep EHR imputers that have been incorporated into the \texttt{PyPOTS} library, ensuring standardised evaluation and experimental control.
\end{enumerate}
In Table \ref{tab:overview}, we list all deep imputation methods surveyed along with their basic component (architecture, framework) inductive bias. In Table \ref{tab:higher}, we delve into the higher-level biases that form due to design modifications applied to the components to further improve imputation performance. In this table, we list the types of missingness based on what has been reported in the original papers. Below is the discussion of the design principles of each of the models we have reviewed: 

\subsubsection{Models Based on Modified Architectures}
Modified RNN-based imputers extend the basic RNN short-term temporal dependency bias by incorporating specialised mechanisms to address different EHR data characteristics.  \textbf{GRUD} \cite{che2018recurrent} aims to address the issue of irregular sampling by embedding an exponential decay assumption where past observations lose relevance at a fixed rate. The decay assumption also aligns the model with the recording patterns of high-frequency variables where both natural physiological decay and equipment-based MNAR patterns (e.g., disconnections) occur. However, GRUD does not explicitly model cross-sectional correlations, which is modeled in \textbf{BRITS} \cite{cao2018brits} via an additional fully-connected layer. BRITS further enhances the robustness of its temporal modeling by utilising bidirectional dynamics. Despite being one of the earliest EHR deep imputers, BRITS' ability to robustly capture temporal and cross-sectional correlations have kept it a popular model today. \textbf{MRNN} \cite{yoon2017multi} aims to address temporal complexity and irregular sampling by incorporating auxiliary variables to capture multi-resolution dependencies, acknowledging varying temporal granularities from minute-level vitals to daily assessments. While this approach improves temporal dependency modeling for with a clinical category (labs, medications, procedures), it can oversimplify the cross-sectional correlations between modalities.

The rise of the Transformer architecture has led to the development of models aiming to take advantage of its global perspective to identify subtle, yet clinically significant patterns that might be overlooked by models with a narrower focus. To preserve the strict sequential integrity that defines clinical time-series \cite{zeng2023transformers, chen2023tsmixer} and to capture equally useful short-term temporal associations \cite{du2020multivariate}, these models use different bespoke modifications. \textbf{GLIMA} \cite{suo2020glima} incorporates both intra-sequence and inter-sequence attention to capture local and global dependencies, \textbf{MTSIT} \cite{yildiz2022multivariate} introduces a modified attention mechanism incorporating temporal positional encoding to preserve sequential information. Finally, \textbf{SAITS} \cite{du2023saits} employs a dual-view self-attention mechanism, where the first view focuses on temporal dependencies within each variable while the second view captures spatio-temporal cross-variable interactions. 


\textbf{TiledCNN} \cite{wang2015imaging} and \textbf{TimesNet} \cite{wu2023timesnet} transform time series data into two-dimensional representations, enabling simultaneous extraction of spatial and temporal features through CNN architecture. This approach uncovers patterns that span multiple time points and variables within physiological signals, such as concurrent analysis of ECG readings, respiratory rates, and oxygen saturation levels \cite{zhao2023deep}. However, these models introduce additional complexity in transforming and interpreting time series in 2D spaces, potentially obscuring the temporal sequence and causal relationships inherent in the data \cite{tang2020rethinking}. TSI-GNN \cite{gordon2021tsi} extends GNNs to represent EHR missingness using a bipartite graph structure that captures spatiotemporal dependencies, where temporal edges connect the same variable across consecutive timestamps while feature edges link different variables within the same timestamp. Although the approach can theoretically capture EHR characteristics, it faces challenges in aligning the static nature of graphs with the dynamic nature of medical time series \cite{jin2023survey}. Additionally, constructing and interpreting these graph models requires significant computational resources and domain knowledge, limiting their scalability and practicality \cite{wu2020connecting}.

\begin{table*}[htbp]
\centering
\begin{small}
\caption{Chronological overview of deep learning models specialised in medical time series imputation. \textbf{Component Bias:} the bias induced by the imputer's components - architecture and (optionally) generative framework. }
\resizebox{\textwidth}{!}{
   \begin{tabular}{lccccc}
   \midrule
   \textbf{Model} & \textbf{Year} & \textbf{Architecture} & \textbf{Component} & \textbf{Uncertainty} \\ 
       \textbf{} & \textbf{} & \textbf{} & \textbf{Bias} & \textbf{Quantification} \\ 
       \midrule    
\textbf{Modified  Single Architectures}& & & & \\
   
   MRNN \cite{yoon2017multi}& 2017  & RNN   & Sequential &  \ding{56} \\ 
   GRUD \cite{che2018recurrent}& 2018  & RNN   &  Sequential & \ding{56} \\ 
   BRITS \cite{cao2018brits}& 2018  & RNN    & Sequential & \ding{56} \\
       \midrule
   
   GLIMA \cite{suo2020glima} & 2020  & Attention &  Globality & \ding{56} \\ 

   MTSIT \cite{yildiz2022multivariate} & 2022  & Attention & Globality & \ding{56} \\ 
   SAITS \cite{du2023saits}& 2023  & Attention & Globality & \ding{56} \\
       \midrule 
   Tiled CNN \cite{wang2015imaging}& 2015  & CNN   & Locality & \ding{56} \\

   TimesNet \cite{wu2023timesnet}&   2023& CNN & Locality & \ding{56} \\ 
    \midrule 
   
   TSI-GNN \cite{gordon2021tsi} & 2021  & GNN   &  Relational & \ding{56} \\

    \midrule 
\textbf{VAEs}& & & & \\
   MIWAE \cite{mattei2019miwae}& 2019  & CNN   & Locality, Stochasticity & \ding{52} \\ 
   GP-VAE \cite{fortuin2020gp}& 2020  & CNN   & Locality, Stochasticity & \ding{52} \\ 
   HI-VAE \cite{nazabal2020handling} & 2020  & MLP   &  Stochasticity & \ding{52} \\ 
      V-RIN \cite{mulyadi2021uncertainty}& 2021  & RNN   &  Sequential, Stochasticity & \ding{52} \\ 
   Shi-VAE \cite{barrejon2021medical} & 2022  & RNN   &  Sequential, Stochasticity & \ding{52} \\ 
   supnot-MIWAE \cite{kim2023probabilistic} & 2023  & CNN, Attention & Locality, Globality, Stochasticity & \ding{52} \\
    \midrule 
        \textbf{MDNs}& & & & \\
   CDNet \cite{liu2022compound}  & 2022  & RNN   &  Sequential, Mixture & \ding{52} \\
    \midrule 
        \textbf{GAN}& & & & \\
    
   VIGAN \cite{shang2017vigan} & 2017  & CNN   &   Locality, Adversariality & \ding{56} \\ 
   GRUI-GAN \cite{luo2018multivariate} & 2018  & RNN   & Sequential, Adversariality & \ding{56} \\ 
   $E^2$GAN \cite{luo2019e2gan} & 2019  & RNN   &   Sequential, Adversariality & \ding{56} \\ 
   NAOMI  \cite{liu2019naomi} & 2019 & MLP  & Adversariality & \ding{56} \\
   US-GAN \cite{miao2021generative} & 2021  & RNN   &   Sequential, Adversariality & \ding{56} \\ 
  Sim-GAN \cite{pati2022missing} & 2022  & CNN   &   Locality, Adversariality & \ding{56} \\
    \midrule 
        \textbf{Diffusion}& & & & \\

   CSDI \cite{tashiro2021csdi}  & 2021  & Attention & Globality, Gradualism & \ding{52} \\ 
   SSSD \cite{alcaraz2022diffusion}  & 2023  & CNN   & Locality, Gradualism & \ding{52} \\ 
   CSBI \cite{chen2023provably} & 2023  & CNN, Attention &  Locality, Globality, Gradualism & \ding{52} \\ 
   DA-TASWDM \cite{xu2023density} & 2023  & Attention &  Globality, Gradualism & \ding{56} \\
    \midrule 
        \textbf{Neural ODEs}& & & & \\
   CRU \cite{schirmer2022modeling}   & 2022  & RNN   & Sequential, Continuity & \ding{52} \\
   CSDE \cite{park2021neural}  & 2022  & MLP   & Continuity & \ding{52} \\
   \bottomrule
   \end{tabular}
}
\label{tab:overview}
\end{small}
\end{table*}

\subsubsection{Models with a Generative Component}

These models operate by generating a latent representation of the EHR time-series in the aim of achieving one which captures the different EHR data properties enabling robust imputation.  \textbf{MIWAE}\cite{mattei2019miwae}, \textbf{GP-VAE }\cite{fortuin2020gp} , \textbf{HI-VAE} \cite{nazabal2020handling}, \textbf{V-RIN} \cite{mulyadi2021uncertainty},  \textbf{Shi-VAE }\cite{barrejon2021medical} and \textbf{Supnot-MIVAE }\cite{kim2023probabilistic}  adopt the VAE framework to map diverse data types and modalities. However, the success of a VAE model depends on its ability to create meaningful latent representations that align with their assumed distribution- a challenge with complex EHR data with numerous clinical subtleties. HI-VAE \cite{nazabal2020handling},  MIWAE \cite{mattei2019miwae} and GP-VAE \cite{fortuin2020gp}, attempt to use architectural bias to improve Gaussian-based predictions, but their performance has been reported to plummet with heterogeneity in observations and extended missingness typical of EHRs \cite{zhang2023comprehensive}. V-RIN \cite{mulyadi2021uncertainty} incorporates an uncertainty-aware Gated Recurrent Unit (GRU) to blend temporal dynamics with the Gaussian imputations. Supnot-MIVAE \cite{kim2023probabilistic} extends this approach by introducing an additional classifier to refine the evidence lower bounds, enhancing imputation accuracy, while Shi-VAE \cite{barrejon2021medical} further expands these capabilities by including LSTMs for better temporal structure handling. While these hybrid VAE models bypass the distribution problem by incorporating temporal dynamics, they face significant challenges in producing clinically relevant outputs. Furthermore, the computational intensity for training VAEs, especially when integrating temporal dynamics, remains a barrier to their wide adoption for large medical datasets. 

MDN deep imputers generate complex, non-Gaussian probabilistic distributions capturing multimodal clinical measurements. However, determining optimal mixture components remains challenging given EHR heterogeneity. \textbf{CDNET} \cite{liu2022compound} addresses this through a compound architecture integrating GRUs and Regularised Attention Networks, enabling simultaneous modelling of temporal dependencies and feature distributions.  This setup allows for robust handling of irregular sampling patterns whilst capturing the underlying multimodal distributions of clinical variables. While promising in theory, MDN models face implementation challenges due to computational complexity and difficulty in optimising parameters whilst maintaining clinical relevance \cite{zhuang2023mixture}.

\textbf{GRUI-GAN} \cite{luo2018multivariate}, \textbf{$E^2$GAN} \cite{luo2019e2gan}, \textbf{NAOMI} \cite{liu2019naomi} and \textbf{US-GAN} \cite{miao2021generative} leverage GAN's adversarial process to generate realistic synthetic EHR time-series through adversarial training, potentially handling irregular sampling and temporal dependencies in EHR data. In order to achieve stable adversarial training dynamics, GRUI-GAN \cite{luo2018multivariate} incorporates a modified GRUs in both the generator and discriminator, but faces difficulties in optimising the noise vector for generation. $E^2$GAN \cite{luo2019e2gan} tackles this limitation by incorporating a denoising autoencoder structure, while NAOMI \cite{liu2019naomi} introduces a non-autoregressive approach to minimise error accumulation in extended sequences. While these hybrid GAN models demonstrate promising capabilities in generating synthetic EHR data, they face significant challenges in ensuring clinical reliability and avoiding mode collapse.  US-GAN \cite{miao2021generative} addresses these issues by implementing a temporal reminder matrix and additional classification layers. The wide adoption of GAN-based models is hindered by their inability quantify confidence, coupled with training instabilitie.

Diffusion-based deep imputers operate by gradually denoising data through an iterative process, aiming to learn the irregular sampling patterns and temporal dependencies through reverse diffusion steps. The success of a diffusion model depends on achieving efficient computation while maintaining accurate temporal modeling. \textbf{CSDI} \cite{tashiro2021csdi} attempts to address this through transformer architectures but faces scalability issues due to quadratic complexity \cite{shen2023non}. \textbf{SSSD} \cite{alcaraz2022diffusion} tackles this limitation using structured state-space models, while \textbf{CSBI} \cite{chen2023provably} and \textbf{DA-TASWDM }\cite{xu2023density} further enhance efficiency by integrating spatio-temporal dependencies and dynamic temporal relationships. While these models demonstrate promising capabilities in handling temporal dependencies, their high computational remains a barrier. 

Finally, \textbf{CRU} \cite{schirmer2022modeling} and \textbf{CSDE} \cite{van1976stochastic} generate continuous data transitions through ODE-learned functions \cite{park2021neural}, addressing the challenge of irregular EHR sampling \cite{rubanova2019latent}. CRU \cite{schirmer2022modeling} employs a linear stochastic differential equation (SDE) \cite{van1976stochastic} within a latent space structure, integrating continuous-discrete Kalman filters with medical time series analysis. CSDE \cite{park2021neural} introduces a probabilistic framework that enhances traditional dynamic models through Markov dynamic programming \cite{howard1960dynamic} and multi-conditional forward-backward losses, enabling robust training and theoretical optimality. However, learning stable dynamics functions remains challenging with sparse and irregular EHR data, and like diffusion models, the computational complexity of neural ODEs continues to be a significant barrier.

\begin{table*}[htbp]
\centering
\caption{Breakdown of the temporal and cross-sectional dependencies and reported missing data strategies of deep EHR imputers.}
\resizebox{\textwidth}{!}{
\begin{tabular}{lcccc}
\midrule
\textbf{Model} & \textbf{Temporal} & \textbf{Spatial} & \textbf{Sparsity} & \textbf{Missingness} \\
\textbf{} & \textbf{Dependencies} & \textbf{Dependencies} & \textbf{Handling} & \textbf{Mechanisms} \\
\midrule
\textbf{Modified Architectures}\\
MRNN & Hierarchical & Weak cross-sectional & Multiple sampling frequencies & MAR \\ 

GRUD & Short-term with decay & None & Irregular sampling via decay & MCAR, MNAR \\ 

BRITS & Bidirectional & Cross-sectional via FC & No explicit handling & MAR \\
\midrule

TimesNet & Local via 2D encoding & Local via convolution & Regular grid assumption & MAR \\ 

Tiled CNN & Multi-scale temporal & Via Gramian fields & Regular grid assumption & MAR \\
\midrule 

GLIMA & Global via attention & Global attention & Attention-based interpolation & MCAR, MAR \\ 

MTSIT & Global with locality & Global attention & Position-based attention & MCAR, MAR \\ 

SAITS & Global with local refinement & Global attention & Self-attention masking & MCAR \\
\midrule 
\textbf{GNNs}\\

TSI-GNN & Graph-structured & Bipartite graph & Graph-based interpolation & MCAR \\
\midrule 
\textbf{VAEs}\\

MIWAE & Latent encoding & Latent space & Probabilistic sampling & MCAR, MAR \\ 

GP-VAE & Gaussian process & Latent space & Gaussian process interpolation & MCAR, MAR, MNAR \\ 

V-RIN & GRU dynamics & Latent space & GRU-based handling & MCAR, MAR \\ 

HI-VAE & Latent encoding & Latent space & Probabilistic encoding & MCAR \\ 

Shi-VAE & LSTM dynamics & Latent space & LSTM-based handling & MAR \\ 

supnot-MIWAE & Local-global combined & Attention-based & Multi-scale handling & MNAR \\
\midrule 
\textbf{MDNs}\\

CDNet & GRU-based & Attention-based & Mixture-based interpolation & - \\
\midrule 
\textbf{GAN}\\

VIGAN & Local adversarial & CNN-based & No explicit handling & - \\ 

GRUI-GAN & Irregular modeling & GRU-based & GRU-based adaptation & - \\ 

$E^2$GAN & Denoising sequence & GRU-based & Denoising-based & - \\ 
NAOMI &  Multiresolution &  Multiresolution & & \\
US-GAN & Reminder matrix & RNN-based & Temporal reminder & MCAR \\ 

Sim-GAN & Local patterns & CNN-based & No explicit handling & - \\
\midrule 
\textbf{Diffusion}\\

CSDI & Gradual denoising & Conditional & Progressive refinement & MCAR, MAR, MNAR \\ 

SSSD & State-space evolution & Local structure & State-space modeling & MCAR, MAR, MNAR \\ 

CSBI & Bridge-based & Local-global & Bridge-based refinement & MAR \\ 

DA-TASWDM & Dynamic attention & Global attention & Attention-based handling & MAR \\
\midrule 
\textbf{Neural ODEs}\\

CRU & Probabilistic states & None & ODE-based interpolation & MAR \\

CSDE & Markov dynamics & None & ODE-based interpolation & - \\
\bottomrule
\end{tabular}
}
\label{tab:higher}
\end{table*}

\section{Model Evaluation: Gaps \& Tradeoffs}

Having established the theoretical foundations of deep EHR imputers, we now examine their practical utility through experimental evaluation. We investigate two key aspects that directly impact the usability of deep imputers formulated as two key objectives:

\begin{enumerate}  [wide, label=\textbf{\arabic*}), labelwidth=!]
\item To quantify the relationship between model complexity and imputation performance, examining whether increasingly sophisticated architectures translate to proportional improvements in accuracy.
\item To assess the effectiveness of current evaluation and benchmarking practices, particularly focusing on how well experimental frameworks align with real-world missingness patterns in clinical data and measure the sensitivity of different deep imputers to increased difficulty in experimental settings. 
\end{enumerate}

\subsection{Current Evaluation Practices \& Limitations}
The evaluation of deep imputation models presents unique challenges due to the inherent nature of missing data - we cannot directly measure accuracy on truly missing values. Instead, evaluation relies on simulating missing data conditions in controlled settings. The dominant approach to address this is through \emph{masking}—a technique where certain data points are deliberately designated as missing during training and evaluation. Masking provides a controlled way to test how an algorithm handles incomplete datasets and is thus essential for performance evaluation.

However, our current evaluation paradigm faces several critical limitations. First, the simulation of missing data must align with the theoretical assumptions and capabilities of the models being tested. Our review of the literature reveals significant discrepancies between how models are evaluated and the missingness patterns they are designed to handle. Second, there is considerable variation in how different models implement and report their evaluation procedures, making direct comparisons between models challenging. Finally, the preprocessing steps and experimental design choices that influence model performance are often underreported or inconsistently applied across studies. These limitations manifest most clearly in the use of masking— while the choice of masking strategy can significantly impact model performance, there is little standardisation in how masking is implemented or reported in the literature. Our systematic review identifies the following key gaps:

\subsubsection{Misalignment with Missingness Assumptions} 
\begin{figure}

\centering
\subfloat[Random Masking]{\includegraphics[width=3cm]{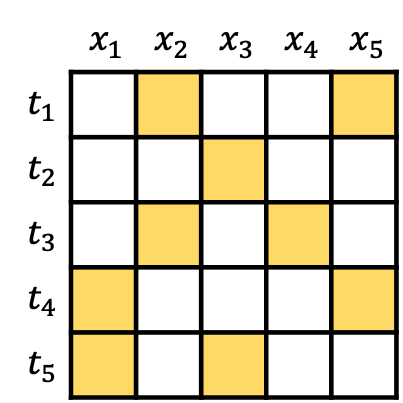}}\hfil
\subfloat[Temporal Masking]{\includegraphics[width=3cm]{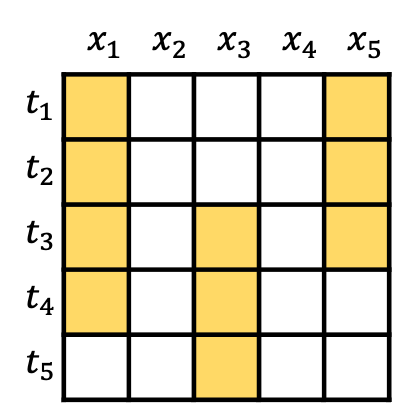}}\hfil

\subfloat[Spatial Masking]{\includegraphics[width=3cm]{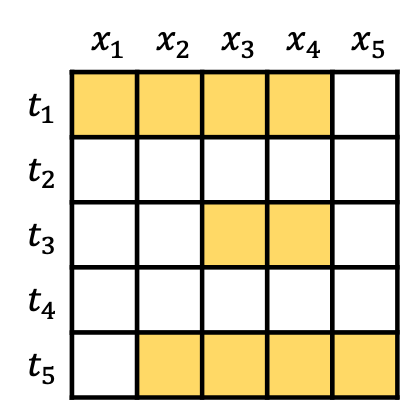}} \hfil 
\subfloat[Block Masking]{\includegraphics[width=3cm]{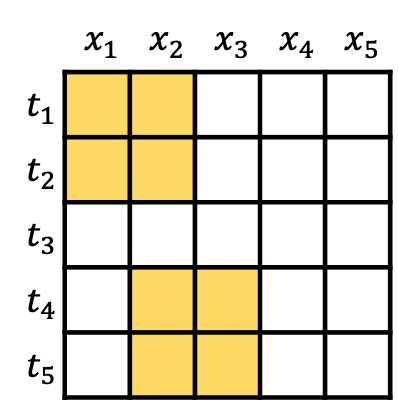}}\hfil   

\subfloat[Augmentation]{\includegraphics[width=3cm]{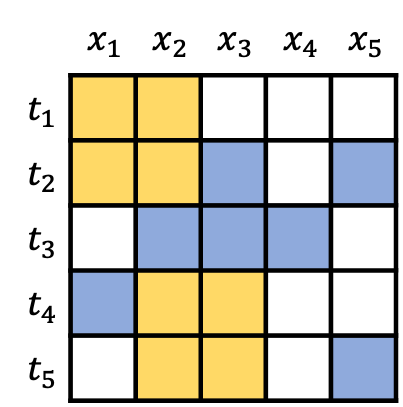}}\hfil
\subfloat[Overlaying]{\includegraphics[width=3cm]{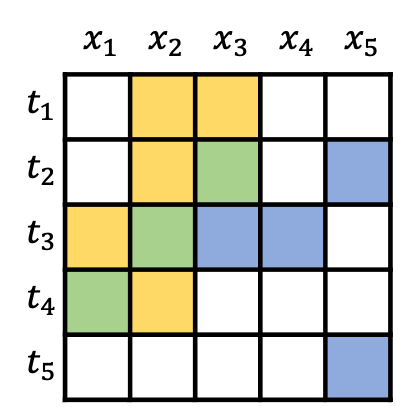}}

\caption{Masking techniques and approaches demonstrated over a time-series of five features ($x_1 - x_5$) and five time points ($t_1 - t_5$): (a) random masking, (b) temporal masking, (c) spatial masking, (d) block masking. The yellow cells indicate those labeled as missing via masking. In (e) augmentation and (f) overlaying, the blue cells indicate cells that are missing within the original data. In (e), the masked (yellow) cells have no overlap with the original missingness in the data. Green: masked data coming from both the original missingness and artificial missingness. In (f), overlaying masks cells from either the original missingness or simulates artificial missingness from non-missing data.} \label{fig:masking}

\end{figure}

As shown in Table \ref{tab:higher}, deep imputers have been designed to recognise different flavours of missingness (MCAR, MAR, MNAR). During experimental evaluation, however, all models shown in Table \ref{tab:higher} use random masking (Figure \ref{fig:masking} (a)) to generate missing datasets, predominantly producing MCAR scenarios and drastically oversimplifying the EHR dependencies as discussed in section \ref{sec:ehrdata}.

Interestingly, the literature contains masking techniques that can capture spatio-temporal missingness patterns\cite{spatiotemporalmasking}, including temporal masking (Figure \ref{fig:masking} (b)), which captures missinngess patterns over time, spatial masking (Figure \ref{fig:masking} (c)), which captures cross-sectional missingness and block masking (Figure \ref{fig:masking} (d)), which combines the two to concurrently capture different flavours of temporal and cross-sectional correlations and dependencies. Despite their direct applicability to biomedical domains, the only examples using spatio-temporal masking of time-series come from the traffic domain \cite{liang2022memory, maskingtraffic2}.

The above problem is exasperated by the lack of information in published work. With the exception of BRITS and CSDI, the use of random masking is not mentioned in the experimental design, and one must examine the accompanying code to discern it. While the use of random masking facilitates model evaluation, it contrasts with the complex and informative MNAR patterns observed in real-world EHRs \cite{garcia2010pattern} which many deep imputers have been designed to address. The discrepancy between the theoretical model and experimental evaluation technique therefore highly undermines a deep imputer's capacity, leaving it under-evaluated. Standardisation tools such as YAIB \cite{van2023yet} and PyPOTS \cite{du2023pypots} offer unified masking functionalities, urging a shift towards more sophisticated and data-driven masking designs.

\subsubsection{Under-reporting of Masking Pipelines} There are significant discrepancies and under-reporting of when masking is introduced during experimental evaluation. Data could be pre-masked before being ingested by the model or masked dynamically during the training phase. Traditional pre-masking methods, while more straightforward, limit the model's training to incomplete datasets, reducing its ability to learn from the entire range of clinical features and associated dependencies. In contrast, adopting in-mini-batch masking strategies promises a more dynamic approach by iteratively masking different subsets of the same dataset across training epochs. However, this approach risks overfitting, as the model may become too focused on the artificial missing patterns and fail to recognise the original data structures. Therefore, the decision of when masking is introduced can have a profound impact on the model's capacity to interpret the diverse missing patterns found in a given dataset \cite{secrets}. Despite the potential impact on the results, this aspect of the experimental design is not reported in most deep imputers discussed in this survey, except for BRITS and GRUD, which mask before training, and CSDI and STAITS, which use in-mini-batch masking during training.

\subsubsection{Overlooked Design Decisions} There are other decisions that highly influence the resulting masked data but are not discussed in most of the deep imputation literature. An important issue is the methodology used to implement masking \cite{m3brits}. Generally, masking can be implemented using \emph{overlaying }\cite{du2023pypots} or \emph{augmenting }\cite{choi2023rdis} as shown in Figures \ref{fig:masking} e-f. Overlaying involves adding artificial missingness in addition to the original missingness the dataset contains, while augmentations only mask complete data, separating the artificial missingness generated from the original missingness. Although overlaying exposes the model to a broader array of missing data scenarios leading to more robust training and effective imputation strategies, it requires complex evaluation processes and increases the risk of overfitting. On the other hand, augmenting simplifies the model's learning process by allowing it to learn from the artificially introduced missingness without interference from the original missing patterns, but may not fully equip the model to handle the real missingness patterns of the data.  It is unclear how any of the deep imputers implement masking, creating a big gap in our understanding of the rigour of the evaluation techniques, especially in models designed to accommodate non-random EHR missingness.


\subsection{Experimental Details}


Our experimental evaluation is designed to satisfy our evaluation objectives stated at the beginning of this section and is performed using the \texttt{PyPOTS} ecosystem \cite{du2023pypots}\footnote{https://pypots.com/about/}, a unified framework for time series imputation. \texttt{PyPOTS} is an openly-available toolkit that provides standardised access to the imputation pipeline, from data preprocessing (including masking) to model evaluation, and houses widely-used benchmarks in its data warehouse.  \texttt{PyPOTS} is the only available platform where benchmarking can be performed on a unified environment using controlled masking functionalities. Using \texttt{PyPOTS}, we are able to compare models using \textbf{a)} vigorously tested and approved implementations of the models evaluated, \textbf{b) }a standardised  implementation environment providing full control of all aspects of the imputation pipeline, including data preprocessing, hyperparameter tuning and masking, and \textbf{c) }an evaluation environment which ensures transparent experimental conditions - key limitations in previous comparative studies where different evaluation frameworks and preprocessing steps have made direct comparisons challenging.

\subsubsection{Dataset Characteristics} We perform our evaluation using the widely-known Physionet Cardiology Challenge 2012 benchmarking dataset (PhysioNet 2012) \cite{physionetchallenge}, which is housed in \texttt{PyPOTS}\footnote{https://github.com/WenjieDu/TSDB}. Physionet 2012 records the first 48 hours of ICU admission for 12,000 ICU patient records from the MIMIC III Clinical database \cite{johnson2016mimic}. The dataset comprises 35 features consisting of vital signs and laboratory tests with varying measurement frequencies and contains a high missingness rate of 79.3\%, consistent with the recording patterns of the measured variables. The missingness rates and density distributions of the Physionet 2012 dataset are provided in the supplementary material as Figure S.1 and S.2 respectively. 

\subsubsection{Models Evaluated} 
\texttt{PyPOTS} incorporates 41 neural network models spanning imputation and downstream tasks. Of these, 8 are deep EHR imputers (Tables \ref{tab:overview} and \ref{tab:higher}) and form the basis of our experimental evaluation. The \texttt{PyPOTS} implementation spans sequential models (BRITS, GRUD, MRNN), convolutional models (TimesNet), attention-based (SAITS), diffusion-based (CSDI), variational (GP-VAE), and adversarial approaches (US-GAN). While RNN-based architectures, being the earliest deep imputers, are well-represented in \texttt{PyPOTS}, the inclusion of highly-performing models using more recent approaches such as Neural ODEs into the package is ongoing work pending code verification and compatibility testing. Nevertheless, this diverse set of architectures enables broad evaluation across different modeling paradigms, allowing us to assess how theoretical advantages translate to practical performance.

\subsubsection{Experiment Design \& Implementation Details}

Relying on \texttt{PyPOTS}' \texttt{PyGrinder}\footnote{https://pypots.com/ecosystem/\#PyGrinder} library to simulate different masking conditions and  \texttt{BenchPOTS}\footnote{https://github.com/WenjieDu/BenchPOTS} library control aspects of benchmarking pipeline, we devise four experiments designed to satisfy our evaluation objectives: 

\begin{enumerate} [wide, label=\textbf{\arabic*}), labelwidth=!]

\item \textbf{Performance-Complexity Tradeoffs:} Examines the relationship between model complexity and imputation accuracy by analysing performance against model size (parameter count), runtime efficiency, and theoretical complexity to determine whether architectural sophistication translates to proportional gains in accuracy.

\item \textbf{Impact of Missingness Mechanisms:} Assesses model robustness and performance variations across different missingness patterns of point-based masking, temporal masking, and block masking.

\item \textbf{Masking Implementation Strategies:} Examines how the timing of masking affects model learning by comparing dynamic in-minibatch masking during training against static masking before training.

\item \textbf{Masking Design Impact:} Investigates how different masking approaches affect model performance by comparing augmentation masking against overlay masking to understand whether simplified masking leads to overoptimistic performance estimates.

\end{enumerate}
In all experiments, we mask 10\% of the data as ground truths. Because the Physionet already contains a high proportion of missing values ($\sim 80\%$), we did not increase the missingness rates in our experiments. 

All experiments were conducted on a machine equipped with an NVIDIA A100 GPU, 80GB of RAM. Each model was trained with its recommended hyperparameters using \texttt{PyGrinder} and tuned through 5-fold cross-validation. The preprocessing scripts, model configurations, and training code are publicly available on \texttt{GitHub}\footnote{https://github.com/LinglongQian/Mask\_rethinking}, promoting transparency and reproducibility for the research community.

Mode performance was compared using the mean absolute error (MAE) and mean squared error (MSE) for added sensitivity to larger errors. Model size was measured using the number of parameters and running time was measured in hours. Collectively, these metrics provide coverage of imputation accuracy and computational efficiency of the model.

\subsection{Experimental Results: Performance Efficiency Tradeoffs }

\begin{figure}[!htbp]

\includegraphics[width=\linewidth]{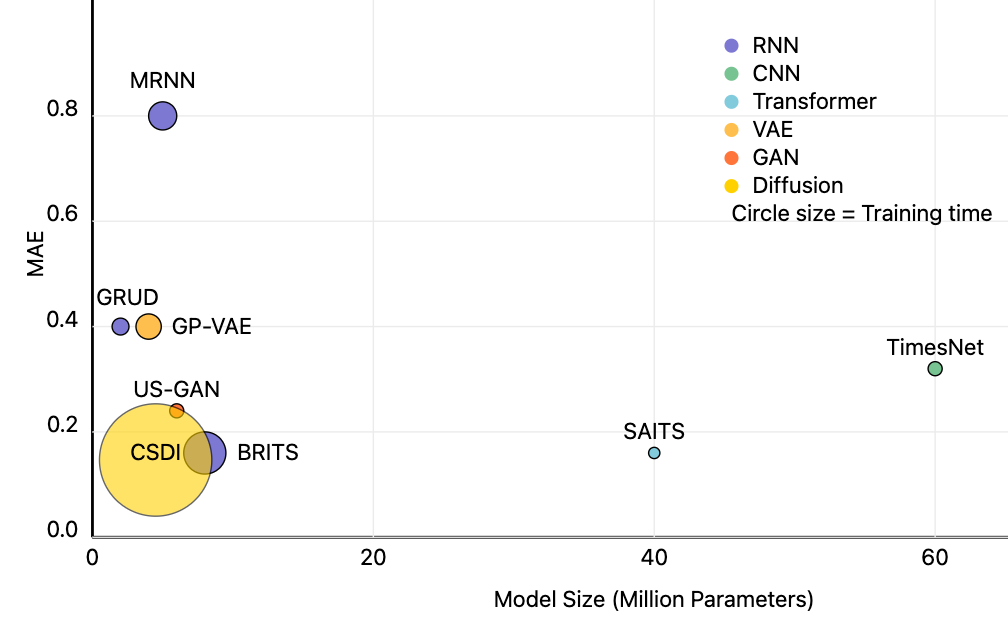}
\caption{Perforamnce Efficiency of the eight models.} 
\label{fig:performanceefficiency}
\end{figure}

The performance reported by the 8 models using 10\% point missingness is shown in Table \ref{tab:physionet_complexity}, along with the theoretical training complexity of the models derived from the original papers. Figure \ref{fig:performanceefficiency} shows the results of comparing the models' MAE with the number of parameters and actual training time (measured in hours). Examining the relationship between model complexity and imputation performance reveals patterns that challenge assumptions about architectural scaling. 

\begin{table}[htbp]
\centering
\setlength{\tabcolsep}{3pt}
\caption{Performance and Complexity Analysis of Deep Imputation Models on PhysioNet 2012:  \textbf{MAE}: Mean Absolute Error; \textbf{MSE}: Mean Squared Error; \textbf{Training Time Complexity}: The computational complexity of one complete training iteration (forward pass, loss computation, backward pass, and parameter updates) as a function of input dimensions.; \textbf{Training Space Complexity:} Memory needed during training (including gradients, intermediate activations, batches) \textbf{N: }sequence length; \textbf{T: }number of features/variables; \textbf{S:} number of diffusion steps (for CSDI)}
\label{tab:physionet_complexity}
\begin{small}
\begin{tabular}{lcccccc}
\toprule

\textbf{Model} & \textbf{MAE} & \textbf{MSE} & \textbf{Time} & \textbf{Space} \\
& & & \textbf{Complexity} & \textbf{Complexity} \\
\midrule
BRITS & 0.297 (0.001) & 0.338 (0.001) & $O(NT)$ & $O(2NT)$ \\
MRNN & 0.708 (0.029) & 0.921 (0.044) & $O(NT)$ & $O(NT)$ \\
GRUD & 0.450 (0.004) & 0.492 (0.006) & $O(NT)$ & $O(NT)$ \\
TimesNet & 0.353 (0.003) & 0.355 (0.004) & $O(NT)$ & $O(NT)$ \\
SAITS & 0.257 (0.019) & 0.276 (0.018) & $O(N^2T)$ & $O(N^2T)$ \\
CSDI & 0.252 (0.004) & 0.313 (0.039) & $O(SN^2T)$ & $O(N^2T)$ \\
GP-VAE & 0.445 (0.006) & 0.499 (0.011) & $O(N^3)$ & $O(N^2)$ \\
US-GAN & 0.310 (0.003) & 0.318 (0.005) & $O(NT)$ & $O(NT)$ \\
\bottomrule

\end{tabular}
\end{small}
\end{table}

CNN-based TimesNet, despite having the largest parameter count (64.9 million parameters), achieves only moderate performance (MAE 0.353), underperforming simpler architectures (BRITS and US-GAN). In contrast, the transformer-based SAITS achieves excellent performance (MAE 0.257) with 44.3 million parameters and surprisingly fast training time (1.02 hours), despite its quadratic complexity. As expected, the single-distribution bias of GP-VAE shows relatively lower performance (MAE 0.445) despite its elegant theoretical foundation. Moreover, despite having 0.5 million parameters, its training time is of 10.59 hours is comparable to that the simpler and poorly-performing MRNN model (MAE: 0.708, number of parameters: 1.6 millions, training time: 10.93 hours). In comparison, GRUD has a comparable MAE to GP-VAE, but with the advantage of a small number of parameters (0.1 million parameters) and faster training (4.78 hours). 

CSDI exemplifies how \textit{architectural innovation can outweigh raw model size}, achieving the best accuracy (MAE 0.252) with only 1.5M parameters. However, this comes with a significant computational cost - 491 hours of training time compared to SAITS's 1.02 hours. This dramatic difference in computational requirements (nearly 500x) highlights the complex trade-offs in model selection and is in line with the fundamental nature of diffusion models, which require multiple forward passes and iterative denoising during training, which is reflected in their time complexity of $O(SN^2T)$, where S is the number of diffusion steps. 

BRITS presents an interesting case in the complexity-performance trade-off: with 9.1M parameters, it achieves strong performance (MAE 0.297), only being outperformed by SAITS and CSDI (much larger models). However, BRITS' training time is significantly large, especially given its relatively modest size (20.06 hours). This apparent inefficiency in RNN-based models like BRITS is a result of their sequential computations - RNNs process data points sequentially, making them inherently difficult to parallelize. While their theoretical complexity is linear ($O(NT)$), each time step must wait for the completion of previous steps.  This suggests that theoretical complexity alone does not determine practical efficiency, and architectural innovations in newer models can overcome their higher computational complexity through better parallelization and more efficient parameter utilization.


\subsection{Experimental Results: The Effect of Missingness Mechanisms}

\begin{figure}[!htbp]

\includegraphics[width=\linewidth]{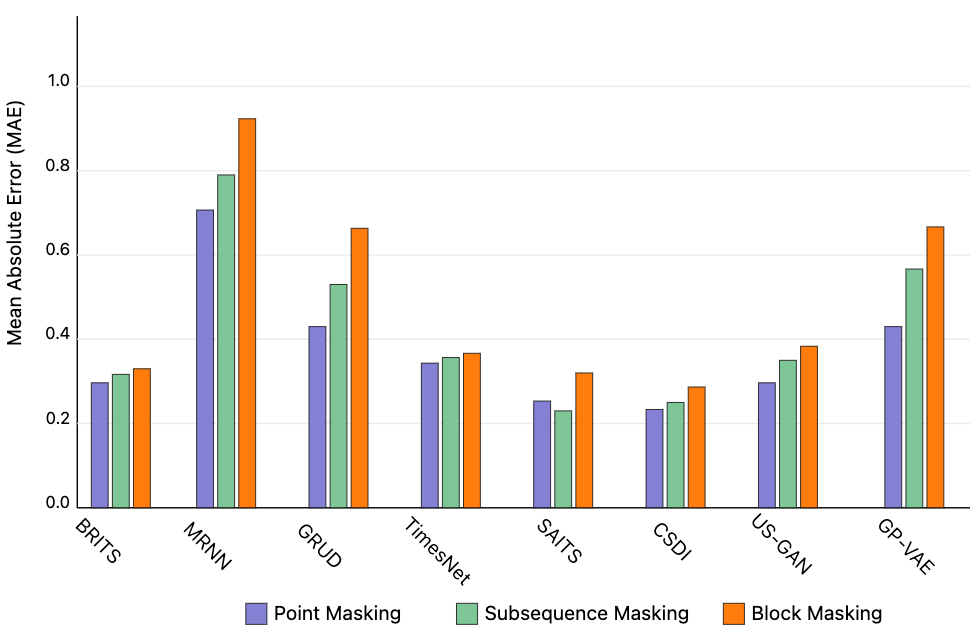}
\caption{The Effect of different masking strategies on  model performance measured in MAE.} 
\label{fig:pointblocksubsequence}
\end{figure}

This experiment assessed the effect of different masking techniques (point, subsequence or block-based masking) on model performance. The results shown in Figure \ref{fig:pointblocksubsequence} reveal that performance progressively degrades as masking patterns become more complex, with block masking consistently yielding the highest MAE scores. This general trend aligns with the increasing complexity of missingness patterns: point masking represents simple MCAR scenarios, subsequence masking captures time-dependent patterns, and block masking introduces both temporal and cross-sectional dependencies. Our results suggest that while no model is immune to the challenges of complex missingness patterns, architectures incorporating dedicated components to capture short and long temporal dependencies and cross-sectional dynamics demonstrate better stability than those focused on local patterns or specific distributional assumptions. Specifically:

\begin{table*}[tbp]
\centering
\caption{Performances with different imputation methods on Physionet 2012 dataset. \textbf{NBM:} Normalisation Before Masking; \textbf{NAM: }Normalisation After Masking. Baselines included: \textbf{LOCF}: Last Observation Carried Forward, \textbf{Median} and \textbf{Mean}. }
\resizebox{\textwidth}{!}{
    \begin{tabular}{l|r|cc|cc|cc}
    \toprule
    \multirow{2}[4]{*}{\textbf{Model}} & \multirow{2}[4]{*}{\textbf{Size}} & \multicolumn{2}{c|}{\textbf{Augmentation Mini-Batch Mask NBM}} & \multicolumn{2}{c|}{\textbf{Augmentation Pre-Mask NBM}} & \multicolumn{2}{c}{\textbf{Augmentation Pre-Mask NAM}} \\
\cmidrule{3-8}          &       & \textbf{MAE} $\downarrow$ & \textbf{MSE} $\downarrow$ & \textbf{MAE} $\downarrow$ & \textbf{MSE} $\downarrow$ & \textbf{MAE} $\downarrow$ & \textbf{MSE} $\downarrow$ \\
    \midrule
    \textbf{SAITS} & 43.6M & \textbf{0.211±0.003} & \textbf{0.268±0.004} & 0.267±0.002 & 0.287±0.001 & 0.267±0.007 & 0.290±0.001 \\
    \textbf{GRUD} & 0.1M & 0.422±0.001 & 0.474±0.003 & 0.483±0.002 & 0.401±0.005 & 0.483±0.002 & 0.403±0.003 \\
    \textbf{TimesNet} & 44.3M & 0.289±0.014 & 0.330±0.019 & 0.288±0.002 & 0.278±0.007 & 0.290±0.002 & 0.279±0.005 \\
    \textbf{CSDI} & 0.3M & 0.239±0.012 & 0.759±0.517 & \textbf{0.237±0.006} & \textbf{0.302±0.057} & \textbf{0.241±0.017} & \textbf{0.430±0.151} \\
    \textbf{GPVAE} & 2.5M & 0.425±0.011 & 0.511±0.017 & 0.399±0.002 & 0.402±0.004 & 0.396±0.001 & 0.401±0.004 \\
    \textbf{US-GAN} & 0.9M & 0.298±0.003 & 0.327±0.005 & 0.294±0.003 & 0.261±0.004 & 0.293±0.002 & 0.261±0.003 \\
    \textbf{BRITS} & 1.3M & 0.263±0.003 & 0.342±0.001 & 0.257±0.001 & 0.256±0.001 & 0.257±0.001 & 0.258±0.001 \\
    \textbf{MRNN} & 0.07M & 0.685±0.002 & 0.935±0.001 & 0.688±0.001 & 0.899±0.001 & 0.690±0.001 & 0.901±0.001 \\
    \textbf{LOCF} & / & 0.411±5.551 & 0.613±0.0 & 0.404±0.0 & 0.506±0.0 & 0.404±0.0 & 0.507±0.0 \\
    \textbf{Median} & / & 0.690±0.0 & 1.049±0.0 & 0.690±0.0 & 1.019±0.0 & 0.691±0.0 & 1.022±0.0 \\
    \textbf{Mean} & / & 0.707±0.0 & 1.022±0.0 & 0.706±0.0 & 0.976±0.0 & 0.706±0.0 & 0.979±1.110 \\
    \midrule
    \multirow{2}[4]{*}{\textbf{Model}} & \multirow{2}[4]{*}{\textbf{Size}} & \multicolumn{2}{c|}{\textbf{Overlay Mini-Batch Mask NBM}} & \multicolumn{2}{c|}{\textbf{Overlay Pre-Mask NBM}} & \multicolumn{2}{c}{\textbf{Overlay Pre-Mask NAM}} \\
\cmidrule{3-8}          &       & \textbf{MAE} $\downarrow$ & \textbf{MSE} $\downarrow$ & \textbf{MAE} $\downarrow$ & \textbf{MSE} $\downarrow$ & \textbf{MAE} $\downarrow$ & \textbf{MSE} $\downarrow$ \\
    \midrule
    \textbf{SAITS} & 43.6M & \textbf{0.206±0.002} & \textbf{0.227±0.005} & 0.274±0.006 & 0.326±0.005 & 0.271±0.006 & 0.325±0.004 \\
    \textbf{GRUD} & 0.1M & 0.419±0.004 & 0.422±0.007 & 0.489±0.002 & 0.436±0.002 & 0.490±0.002 & 0.436±0.003 \\
    \textbf{TimesNet} & 44.3M & 0.273±0.011 & 0.242±0.018 & 0.293±0.003 & 0.290±0.011 & 0.291±0.003 & 0.288±0.007 \\
    \textbf{CSDI} & 0.3M & 0.226±0.010 & 0.279±0.051 & \textbf{0.253±0.005} & \textbf{0.461±0.074} & \textbf{0.239±0.006} & \textbf{0.344±0.109} \\
    \textbf{GPVAE} & 2.5M & 0.427±0.006 & 0.453±0.008 & 0.412±0.007 & 0.484±0.013 & 0.420±0.009 & 0.489±0.008 \\
    \textbf{US-GAN} & 0.9M & 0.295±0.002 & 0.261±0.007 & 0.297±0.002 & 0.284±0.005 & 0.299±0.003 & 0.287±0.004 \\
    \textbf{BRITS} & 1.3M & 0.254±0.001 & 0.265±0.001 & 0.262±0.000 & 0.288±0.003 & 0.263±0.001 & 0.294±0.003 \\
    \textbf{MRNN} & 0.07M & 0.682±0.000 & 0.905±0.001 & 0.685±0.001 & 0.926±0.002 & 0.684±0.001 & 0.923±0.002 \\
    \textbf{LOCF} & / & 0.411±0.0 & 0.532±0.0 & 0.408±0.0 & 0.540±0.0 & 0.409±0.0 & 0.540±0.0 \\
    \textbf{Median} & / & 0.687±0.0 & 1.019±0.0 & 0.686±0.0 & 1.030±0.0 & 0.686±0.0 & 1.030±0.0 \\
    \textbf{Mean} & / & 0.705±0.0 & 0.990±1.110 & 0.702±0.0 & 1.001±0.0 & 0.702±0.0 & 1.000±0.0 \\
    \bottomrule
    \end{tabular}%
}
\label{Physionet_result}
\end{table*}

BRITS, CSDI and TimesNet show minimal degradation (MAE ranges 0.297-0.32, 0.25-0.28 and 0.33-0.36 respectively), validating their designs and demonstrating that their dedicated components effectively capture both temporal and cross-variable interactions even under complex missingness.
SAITS exhibits a unique pattern, being the only model with a lower MAE for sequence masking compared to point masking, demonstrating that its dual-view attention mechanism effectively captures short and long-term temporal dependencies. Both SAITS and US-GAN show moderate degradation to block missingness (MAE ranges 0.24-0.31 and 0.31 to 0.37 respectively), suggesting their dedicated components provide some robustness to complex missingness.
GRUD, MRNN and GP-VAE show the highest sensitivity to masking complexity (MAE ranges: 0.45-0.65,  0.708-0.90 and 0.445-0.65 respectively), suggesting that their design assumptions are not representative of the complexities of EHR data missingnes and confirming known VAE limitations with heterogeneous data.

\subsection{Experimental Results: The Effect of Masking Design Decisions}

This experiments assessed\textbf{ a) }the timing of masking application, either pre-masking the dataset or masking dynamically during training through in-minibatch, \textbf{b) } the masking strategy, either overlying or augmentation, and \textbf{c) }the timing of normalisation with respect to masking (before or after). 

The results are shown in Table \ref{Physionet_result}. SAITS demonstrated superior performance, achieving the lowest MAE of 0.206 under overlay mini-batch mask. Its strong results highlight its effectiveness in handling complex, high-dimensional missingness. CSDI showed MAE as low as 0.226 under augmentation mini-batch mask. However, its sensitivity to masking variability (e.g., MAE 0.253 under overlay pre-masking) highlights potential limitations in dynamic settings. TimesNet showed higher MAE values (e.g., 0.290 under augmentation pre-mask). GRUD, GP-VAE and MRNN underperformed significantly, with MAE exceeding 0.4, reflecting their limitations in capturing high-dimensional dependencies.
 
\section{Discussion}

In this review, we have examined deep learning imputation approaches for EHR data, highlighting the relationship between the inductive biases of these models and the distinct characteristics of medical time-series.  The landscape of deep imputation for electronic health records (EHRs) reveals how fundamental architectural and framework biases synergistically generate higher-level modeling approaches that profoundly shape data reconstruction capabilities. 

Our experimental evaluation highlights the critical role of preprocessing and masking strategies. Our findings emphasise the need for tailored approaches that align model design with EHR data characteristics. By providing detailed evaluations and insights, this work offers a foundation for improving the reliability and robustness of EHR imputers. Future efforts should focus on developing adaptive and hybrid strategies to address the challenges identified, particularly complex missingness patterns and extreme data sparsity.  

\subsection{Key Findings}

\subsubsection{Design Choices Matter}
Neural architectures exhibits inherent inductive preferences that cascade into sophisticated imputation strategies. However, design decisions can complement architectural bias to accommodate data characteristics. Our taxonomy and experimental results show that model sophistication is not directly linked to parameter count, but to the sophistication of bias integration to accommodate the complex features of EHR data. For example, BRITS' thoughtful design transcends RNNs' basic sequential modelling limitations by explicitly incorporating cross-sectional relationships through an additional fully connected layer, demonstrating how carefully engineered architectural modifications can outperform some of the more elaborate models. Similarly, SAITS integrates global attention with local refinement mechanisms,  generating higher-order representations that overcome architectural limitations. 

\subsubsection{Alignment with EHR Characteristics Matters}
The most compelling models emerge when architectural and framework biases are deliberately aligned with clinical data characteristics, and clear deviation leads to suboptimal performance. This is exemplified by MRNN, where its multi-resolution temporal dynamic oversimplifies the cross-sectional relationships between different clinical measurements, and GRUD's exponential decay assumption, though innovative for handling irregular sampling, does not adequately account for the non-linear interdependencies between variables that emerge in complex medical trajectories. Similarly, VAEs are challenged by the misalignment between their distributional assumption and the heterogeneous nature of medical time series. VAEs' limitations may be overcome by MDNs' mixed distributions, but this remains untested by us as \texttt{PyPOTS} does not yet include an MDN model. 

\subsubsection{Benchmarking Design Matters} 
Our experimental evaluation demonstrates that rigorous benchmarking requires careful consideration of both masking strategies and implementation choices. The significant performance variations observed under different masking patterns (point, temporal, and block) reveal that simplistic random masking may lead to over-optimistic performance estimates that don't reflect the real missingness patterns within the data. Furthermore, implementation decisions such as the timing of masking application (pre-mask vs. mini-batch) and masking strategy (overlay vs. augmentation) significantly impact model evaluation, with performance variations of up to 20\% in MAE. These findings emphasise that standardised, comprehensive benchmarking frameworks such as \texttt{PyPOTS} are essential for meaningful model comparison and assessment of practical utility.

\subsection{Open Research Questions}

\subsubsection{The Structure of EHR Missingness}
Our taxonomy and experimental results reveal an important gap in how missingness is conceptualised and evaluated in deep EHR imputers. While Table \ref{tab:higher} shows models claiming to handle different types of missingness (MCAR, MAR, MNAR), our experiments point to a more fundamental issue: the presence of \emph{structured missingness} \cite{mitra2023learning} in clinical data, where the specific modes of data collection cause missing values to exhibit associations and structural patterns. In EHRs, structural missingness naturally arises from the asynchronous and decision-driven nature of healthcare data collection \cite{wahl2018artificial}, reflecting the complex interplay between clinical protocols, resource availability, and care delivery patterns. For instance, vital signs measured at fixed intervals (creating regular gaps), lab tests clustered around clinical events and observations clustered around shift ends. The patterns of structured missingness in EHRs are also shaped by the uneven data distributions of clinical outcomes. Many severe clinical events are rare \cite{rare}. For example, cardiac arrests constitute only 2.3\% of ICU admissions \cite{armstrong2019incidence}, making data samples with a cardiac arrest outcome a minority. This shapes distinct missingness structures: these cases trigger intensive monitoring protocols with frequent measurements, yet their rarity means few complete examples of these measurement patterns exist. This creates a systematic relationship between missingness patterns and clinical severity - the frequency and timing of missing values becomes informative about the underlying patient state.

Our evaluation using different masking patterns (point, temporal, and block) shows that even sophisticated models struggle when missingness patterns become more structured, suggesting current approaches may be oversimplified. These findings highlight that structured missingness extends beyond Rubin's classification of missingness into MCAR, MAR, and MNAR, highlighting the need for new theoretical foundations that better capture the systematic, informative nature of missing patterns in clinical data collection.

\subsubsection{The Challenge of Uncertainty Quantification}
While our taxonomy shows several probabilistic frameworks offer inherent uncertainty estimation, these frameworks rely on distribution-specific assumptions that may struggle to capture the heterogeneous nature of medical time series, while their computational complexity limits practical application. More concerning is that highly performing models such as BRITS and SAITS  are fundamentally deterministic with no inherent capability to communicate imputation confidence. Although emerging approaches like DEARI \cite{deari} and CF-RNN \cite{stankeviciute2021conformal} show promise through post-hoc uncertainty estimation, they remain preliminary solutions. The field needs model-agnostic uncertainty quantification approaches that can adapt to the diverse characteristics of medical time series while maintaining computational efficiency, particularly given the demonstrated importance of uncertainty measures for clinical trust \cite{fortuin2020gp}.

\subsubsection{The Gap Between Models and Domain Knowledge}
Our review highlights a disconnect between computational sophistication and clinical expertise in current deep imputers. While recent work has shown promising directions through the integration of clinical guidelines \cite{csai} and temporal logic \cite{yan2022neuro}, these remain exceptions rather than the norm. Most models in our taxonomy treat medical time series as abstract mathematical constructs, overlooking the rich contextual knowledge embedded in clinical practices. This gap becomes particularly problematic when handling complex distributions and rare events common in medical data, where domain knowledge could help ensure clinically meaningful imputations and downstream results. The challenge extends beyond simple rule integration – models need to model the clinical significance of temporal patterns, ensure physiologically plausible reconstructions, and maintain data integrity across diverse patient populations. Future research must develop systematic approaches to bridge this gap, ensuring imputed values are not just statistically sound but clinically meaningful.



\section*{Acknowledgments}

This paper represents independent research funded by the NIHR Maudsley Biomedical Research Centre at South London and Maudsley NHS Foundation Trust, the EPSRC Centre for Doctoral Training in Data-Driven Health (DRIVE-Health) and King’s College London. LQ is supported by the Kings-China Scholarship Council PhD Scholarship Programme (K-CSC) under Grant CSC202008060096. ZI is supported by Innovate UK under grant 10104845. TW is supported by the NIHR Maudsley Biomedical Research Centre at South London, Maudsley Charity and Early Career Research Award from the Institute of Psychiatry, Psychology \& Neuroscience. ZI and RD were supported by in part by the NIHR Biomedical Research Centre at SLaM, in part by Kings College London, London, U.K., and in part by the NIHR University College London Hospitals Biomedical Research Centre. H.L.E is supported by a Research Fellowship Award from the Department of Medicine at Dalhousie University and DRIVE-Health. RM's research on Structured Missingness has been supported by the Turing-Roche Strategic Partnership. The authors would like to acknowledge KCL CREATE for providing the HPC environment enabling our experimenal work.


\section*{References}

\bibliographystyle{IEEEtran}
\bibliography{references}

\begin{thebibliography}{10}

\bibitem{correlation2}
David~J Albers, Matthew~E Levine, Andrew Stuart, Lena Mamykina, Bruce Gluckman, and George Hripcsak.
\newblock Mechanistic machine learning: how data assimilation leverages physiologic knowledge using bayesian inference to forecast the future, infer the present, and phenotype.
\newblock {\em Journal of the American Medical Informatics Association}, 25(10):1392--1401, 2018.

\bibitem{alcaraz2022diffusion}
Juan~Lopez Alcaraz and Nils Strodthoff.
\newblock Diffusion-based time series imputation and forecasting with structured state space models.
\newblock {\em Transactions on Machine Learning Research}, 2022.

\bibitem{deeppredictionreview2}
Ali Amirahmadi, Mattias Ohlsson, and Kobra Etminani.
\newblock Deep learning prediction models based on ehr trajectories: A systematic review.
\newblock {\em Journal of Biomedical Informatics}, 144:104430, 2023.

\bibitem{armstrong2019incidence}
Richard~A Armstrong, Caroline Kane, Fiona Oglesby, Katie Barnard, Jasmeet Soar, and Matt Thomas.
\newblock The incidence of cardiac arrest in the intensive care unit: A systematic review and meta-analysis.
\newblock {\em Journal of the Intensive Care Society}, 20(2):144--154, 2019.

\bibitem{barrejon2021medical}
Daniel Barrej{\'o}n, Pablo~M Olmos, and Antonio Art{\'e}s-Rodr{\'\i}guez.
\newblock Medical data wrangling with sequential variational autoencoders.
\newblock {\em IEEE Journal of Biomedical and Health Informatics}, 26(6):2737--2745, 2021.

\bibitem{deepsepsis}
Aaron Boussina, Supreeth~P. Shashikumar, Atul Malhotra, Robert~L. Owens, Robert El-Kareh, Christopher~A. Longhurst, Kimberly Quintero, Allison Donahue, Theodore~C. Chan, Shamim Nemati, and Gabriel Wardi.
\newblock Impact of a deep learning sepsis prediction model on quality of care and survival.
\newblock {\em npj Digital Medicine}, 7:14, 2024.

\bibitem{cao2018brits}
Wei Cao, Dong Wang, Jian Li, Hao Zhou, Lei Li, and Yitan Li.
\newblock Brits: Bidirectional recurrent imputation for time series.
\newblock {\em Advances in neural information processing systems}, 31, 2018.

\bibitem{rare}
Adrian~R. Cartus, Elliot~A. Samuels, Magdalena Cerdá, and Brandon D.~L. Marshall.
\newblock Outcome class imbalance and rare events: An underappreciated complication for overdose risk prediction modeling.
\newblock {\em Addiction}, 118(6):1167--1176, June 2023.

\bibitem{che2018recurrent}
Zhengping Che, Sanjay Purushotham, Kyunghyun Cho, David Sontag, and Yan Liu.
\newblock Recurrent neural networks for multivariate time series with missing values.
\newblock {\em Scientific reports}, 8(1):1--12, 2018.

\bibitem{chen2018neural}
Ricky~TQ Chen, Yulia Rubanova, Jesse Bettencourt, and David~K Duvenaud.
\newblock Neural ordinary differential equations.
\newblock {\em Advances in neural information processing systems}, 31, 2018.

\bibitem{chen2023tsmixer}
Si-An Chen, Chun-Liang Li, Nate Yoder, Sercan~O Arik, and Tomas Pfister.
\newblock Tsmixer: An all-mlp architecture for time series forecasting.
\newblock {\em arXiv preprint arXiv:2303.06053}, 2023.

\bibitem{chen2023provably}
Yu~Chen, Wei Deng, Shikai Fang, Fengpei Li, Nicole~Tianjiao Yang, Yikai Zhang, Kashif Rasul, Shandian Zhe, Anderson Schneider, and Yuriy Nevmyvaka.
\newblock Provably convergent schr{\"o}dinger bridge with applications to probabilistic time series imputation.
\newblock In {\em International Conference on Machine Learning}, pages 4485--4513. PMLR, 2023.

\bibitem{choi2023rdis}
Tae-Min Choi, Ji-Su Kang, and Jong-Hwan Kim.
\newblock Rdis: Random drop imputation with self-training for incomplete time series data.
\newblock {\em IEEE Access}, 2023.

\bibitem{creswell2018generative}
Antonia Creswell, Tom White, Vincent Dumoulin, Kai Arulkumaran, Biswa Sengupta, and Anil~A Bharath.
\newblock Generative adversarial networks: An overview.
\newblock {\em IEEE signal processing magazine}, 35(1):53--65, 2018.

\bibitem{mllandscape}
Ritankar Das and David~J. Wales.
\newblock Machine learning landscapes and predictions for patient outcomes.
\newblock {\em Royal Society Open Science}, 4:170175, 2017.

\bibitem{spatiotemporalmasking}
Min Deng, Zide Fan, Qiliang Liu, and Jianya Gong.
\newblock A hybrid method for interpolating missing data in heterogeneous spatio-temporal datasets.
\newblock {\em ISPRS International Journal of Geo-Information}, 5(2), 2016.

\bibitem{du2020multivariate}
Shengdong Du, Tianrui Li, Yan Yang, and Shi-Jinn Horng.
\newblock Multivariate time series forecasting via attention-based encoder--decoder framework.
\newblock {\em Neurocomputing}, 388:269--279, 2020.

\bibitem{du2023pypots}
Wenjie Du.
\newblock Pypots: A python toolbox for data mining on partially-observed time series.
\newblock {\em arXiv preprint arXiv:2305.18811}, 2023.

\bibitem{du2023saits}
Wenjie Du, David C{\^o}t{\'e}, and Yan Liu.
\newblock Saits: Self-attention-based imputation for time series.
\newblock {\em Expert Systems with Applications}, 219:119619, 2023.

\bibitem{tsibench}
Wenjie Du, Jun Wang, Linglong Qian, Yiyuan Yang, Fanxing Liu, Zepu Wang, Zina Ibrahim, Haoxin Liu, Zhiyuan Zhao, Yingjie Zhou, et~al.
\newblock Tsi-bench: Benchmarking time series imputation.
\newblock {\em arXiv preprint arXiv:2406.12747}, 2024.

\bibitem{fortuin2020gp}
Vincent Fortuin, Dmitry Baranchuk, Gunnar R{\"a}tsch, and Stephan Mandt.
\newblock {GP-VAE}: Deep probabilistic time series imputation.
\newblock In {\em International conference on artificial intelligence and statistics}, pages 1651--1661. PMLR, 2020.

\bibitem{garcia2010pattern}
Pedro~J Garc{\'\i}a-Laencina, Jos{\'e}-Luis Sancho-G{\'o}mez, and An{\'\i}bal~R Figueiras-Vidal.
\newblock Pattern classification with missing data: a review.
\newblock {\em Neural Computing and Applications}, 19:263--282, 2010.

\bibitem{gordon2021tsi}
David Gordon, Panayiotis Petousis, Henry Zheng, Davina Zamanzadeh, and Alex~AT Bui.
\newblock Tsi-gnn: Extending graph neural networks to handle missing data in temporal settings.
\newblock {\em Frontiers in big Data}, 4:693869, 2021.

\bibitem{inductivebias}
A~Goyal and Bengiom Y.
\newblock Inductive biases for deep learning of higher-level cognition.
\newblock {\em Proceedings of the Royal Society}, 478:20210068, 2022.

\bibitem{ho2020denoising}
Jonathan Ho, Ajay Jain, and Pieter Abbeel.
\newblock Denoising diffusion probabilistic models.
\newblock {\em Advances in neural information processing systems}, 33:6840--6851, 2020.

\bibitem{howard1960dynamic}
R.A. Howard.
\newblock {\em Dynamic programming and Markov processes}.
\newblock Technology Press of Massachusetts Institute of Technology, 1960.

\bibitem{islam}
Khandaker~Reajul Islam, Johayra Prithula, Jaya Kumar, Toh~Leong Tan, Mamun Bin~Ibne Reaz, Md. Shaheenur~Islam Sumon, and Muhammad E.~H. Chowdhury.
\newblock Machine learning-based early prediction of sepsis using electronic health records: A systematic review.
\newblock {\em Journal of Clinical Medicine}, 12(17), 2023.

\bibitem{jensen2012mining}
Peter~B Jensen, Lars~J Jensen, and S{\o}ren Brunak.
\newblock Mining electronic health records: towards better research applications and clinical care.
\newblock {\em Nature Reviews Genetics}, 13(6):395--405, 2012.

\bibitem{jin2023survey}
Ming Jin, Huan~Yee Koh, Qingsong Wen, Daniele Zambon, Cesare Alippi, Geoffrey~I Webb, Irwin King, and Shirui Pan.
\newblock A survey on graph neural networks for time series: Forecasting, classification, imputation, and anomaly detection.
\newblock {\em arXiv preprint arXiv:2307.03759}, 2023.

\bibitem{johnson2016mimic}
Alistair Johnson and et~al.
\newblock Mimic-iii, a freely accessible critical care database.
\newblock {\em Scientific data}, 3(1):1--9, 2016.

\bibitem{reviewhealthcare}
Maksims Kazijevs and Manar~D. Samad.
\newblock Deep imputation of missing values in time series health data: A review with benchmarking.
\newblock {\em Journal of Biomedical Informatics}, 144:104440, 2023.

\bibitem{kim2023probabilistic}
SeungHyun Kim, Hyunsu Kim, Eunggu Yun, Hwangrae Lee, Jaehun Lee, and Juho Lee.
\newblock Probabilistic imputation for time-series classification with missing data.
\newblock In {\em International Conference on Machine Learning}, pages 16654--16667. PMLR, 2023.

\bibitem{cardiacarrest}
Youngjae Lee, Joon-myoung Kwon, Yeha Lee, Hyunho Park, Hyungchul Cho, and Jinsik Park.
\newblock Deep learning in the medical domain: Predicting cardiac arrest using deep learning.
\newblock {\em Acute and Critical Care}, 33(3):117--120, Aug 2018.

\bibitem{lee2018deep}
Youngjae Lee, Joon-myoung Kwon, Yeha Lee, Hyunho Park, Hyungchul Cho, and Jinsik Park.
\newblock Deep learning in the medical domain: Predicting cardiac arrest using deep learning.
\newblock {\em Acute and Critical Care}, 33(3):117--120, 2018.

\bibitem{mnarnature}
Jiaming Li, Xuejun~S Yan, Dandi Chaudhary, Shenghui You, Xiaoqian Wang, Suresh~K Bhavnani, Luke~V Rasmussen, Yixuan Zhou, and Cui Tao.
\newblock Imputation of missing values for electronic health record laboratory data.
\newblock {\em npj Digital Medicine}, 4(1):147, 2021.

\bibitem{liang2022memory}
Yuebing Liang, Zhan Zhao, and Lijun Sun.
\newblock Memory-augmented dynamic graph convolution networks for traffic data imputation with diverse missing patterns.
\newblock {\em Transportation Research Part C: Emerging Technologies}, 143:103826, 2022.

\bibitem{reviewhealthcare2}
Mingxuan Liu, Siqi Li, Han Yuan, Marcus Eng~Hock Ong, Yilin Ning, Feng Xie, Seyed~Ehsan Saffari, Yuqing Shang, Victor Volovici, Bibhas Chakraborty, and Nan Liu.
\newblock Handling missing values in healthcare data: A systematic review of deep learning-based imputation techniques.
\newblock {\em Artificial Intelligence in Medicine}, 142:102587, 2023.

\bibitem{liu2019naomi}
Yukai Liu, Rose Yu, Stephan Zheng, Eric Zhan, and Yisong Yue.
\newblock Naomi: Non-autoregressive multiresolution sequence imputation.
\newblock {\em Advances in neural information processing systems}, 32, 2019.

\bibitem{liu2022compound}
Yuxi Liu, Shaowen Qin, Zhenhao Zhang, and Wei Shao.
\newblock Compound density networks for risk prediction using electronic health records.
\newblock In {\em 2022 IEEE International Conference on Bioinformatics and Biomedicine (BIBM)}, pages 1078--1085. IEEE, 2022.

\bibitem{luo2018multivariate}
Yonghong Luo, Xiangrui Cai, Ying Zhang, Jun Xu, et~al.
\newblock Multivariate time series imputation with generative adversarial networks.
\newblock {\em Advances in neural information processing systems}, 31, 2018.

\bibitem{luo2019e2gan}
Yonghong Luo, Ying Zhang, Xiangrui Cai, and Xiaojie Yuan.
\newblock E2gan: End-to-end generative adversarial network for multivariate time series imputation.
\newblock In {\em Proceedings of the 28th international joint conference on artificial intelligence}, pages 3094--3100. AAAI Press Palo Alto, CA, USA, 2019.

\bibitem{luo2022evaluating}
Yuan Luo.
\newblock Evaluating the state of the art in missing data imputation for clinical data.
\newblock {\em Briefings in Bioinformatics}, 23(1):bbab489, 2022.

\bibitem{mackay1998introduction}
David~JC MacKay et~al.
\newblock Introduction to gaussian processes.
\newblock {\em NATO ASI series F computer and systems sciences}, 168:133--166, 1998.

\bibitem{mattei2019miwae}
Pierre-Alexandre Mattei and Jes Frellsen.
\newblock Miwae: Deep generative modelling and imputation of incomplete data sets.
\newblock In {\em International conference on machine learning}, pages 4413--4423. PMLR, 2019.

\bibitem{medsker2001recurrent}
Larry~R Medsker and LC~Jain.
\newblock Recurrent neural networks.
\newblock {\em Design and Applications}, 5(64-67):2, 2001.

\bibitem{miao2021generative}
Xiaoye Miao, Yangyang Wu, Jun Wang, Yunjun Gao, Xudong Mao, and Jianwei Yin.
\newblock Generative semi-supervised learning for multivariate time series imputation.
\newblock In {\em Proceedings of the AAAI conference on artificial intelligence}, volume~35, pages 8983--8991, 2021.

\bibitem{complex}
R.~Miotto, F.~Wang, S.~Wang, X.~Jiang, and J.T. Dudley.
\newblock Deep learning for healthcare: review, opportunities and challenges.
\newblock {\em Briefings in Bioinformatics}, 19(6):1236--1246, Nov 2018.

\bibitem{mitra2023learning}
Robin Mitra, Sarah~F McGough, Tapabrata Chakraborti, Chris Holmes, Ryan Copping, Niels Hagenbuch, Stefanie Biedermann, Jack Noonan, Brieuc Lehmann, Aditi Shenvi, et~al.
\newblock Learning from data with structured missingness.
\newblock {\em Nature Machine Intelligence}, 5(1):13--23, 2023.

\bibitem{moskovitch2009medical}
Robert Moskovitch and Yuval Shahar.
\newblock Medical temporal-knowledge discovery via temporal abstraction.
\newblock {\em AMIA Annual Symposium Proceedings}, 2009:452--456, 2009.

\bibitem{moss2017continuous}
Travis~J Moss, Douglas~E Lake, J~Forrest Calland, Kyle~B Enfield, John~B Delos, Karen~D Fairchild, and J~Randall Moorman.
\newblock Continuous vital sign analysis for predicting and preventing noncardiac complications after major surgery.
\newblock {\em American Journal of Physiology-Heart and Circulatory Physiology}, 312(4):H627--H636, 2017.

\bibitem{mulyadi2021uncertainty}
Ahmad~Wisnu Mulyadi, Eunji Jun, and Heung-Il Suk.
\newblock Uncertainty-aware variational-recurrent imputation network for clinical time series.
\newblock {\em IEEE Transactions on Cybernetics}, 52(9):9684--9694, 2021.

\bibitem{nazabal2020handling}
Alfredo Nazabal, Pablo~M Olmos, Zoubin Ghahramani, and Isabel Valera.
\newblock Handling incomplete heterogeneous data using vaes.
\newblock {\em Pattern Recognition}, 107:107501, 2020.

\bibitem{oberdiek2022uqgan}
Philipp Oberdiek, Gernot Fink, and Matthias Rottmann.
\newblock Uqgan: A unified model for uncertainty quantification of deep classifiers trained via conditional gans.
\newblock {\em Advances in Neural Information Processing Systems}, 35:21371--21385, 2022.

\bibitem{park2021neural}
Sung~Woo Park, Kyungjae Lee, and Junseok Kwon.
\newblock Neural markov controlled sde: Stochastic optimization for continuous-time data.
\newblock In {\em International Conference on Learning Representations}, 2021.

\bibitem{pati2022missing}
Soumen~Kumar Pati, Manan~Kumar Gupta, Rinita Shai, Ayan Banerjee, and Arijit Ghosh.
\newblock Missing value estimation of microarray data using sim-gan.
\newblock {\em Knowledge and Information Systems}, 64(10):2661--2687, 2022.

\bibitem{correlation1}
Rimma Pivovarov, David~J Albers, George Hripcsak, Jorge~L Sepulveda, and No{\'e}mie Elhadad.
\newblock Temporal trends of hemoglobin a1c testing.
\newblock {\em Journal of the American Medical Informatics Association}, 21(6):1038--1044, 2014.

\bibitem{pivovarov2014identifying}
Rimma Pivovarov, David~J Albers, Jorge~L Sepulveda, and No{\'e}mie Elhadad.
\newblock Identifying and mitigating biases in ehr laboratory tests.
\newblock {\em Journal of biomedical informatics}, 51:24--34, 2014.

\bibitem{deari}
Linglong Qian, Zina Ibrahim, and Richard Dobson.
\newblock Uncertainty-aware deep attention recurrent neural network for heterogeneous time series imputation.
\newblock {\em arXiv preprint arXiv:2401.02258}, 2024.

\bibitem{secrets}
Linglong Qian, Zina Ibrahim, Wenjie Du, Yiyuan Yang, and Richard~JB Dobson.
\newblock Unveiling the secrets: How masking strategies shape time series imputation.
\newblock {\em arXiv preprint arXiv:2405.17508}, 2024.

\bibitem{csai}
Linglong Qian, Zina Ibrahim, Hugh~Logan Ellis, Ao~Zhang, Yuezhou Zhang, Tao Wang, and Richard Dobson.
\newblock Knowledge enhanced conditional imputation for healthcare time-series, 2024.

\bibitem{m3brits}
Linglong Qian, Zina~M. Ibrahim, Ao~Zhang, and Richard J.~B. Dobson.
\newblock Addressing class imbalance in electronic health records data imputation.
\newblock In {\em Proceedings of the 6th International Workshop on Knowledge Discovery from Healthcare Data co-located with 32nd International Joint Conference on Artificial Intelligence {(IJCAI} 2023)}, volume 3479 of {\em {CEUR} Workshop Proceedings}. CEUR-WS.org, 2023.

\bibitem{glucose}
David Rodbard.
\newblock Continuous glucose monitoring: a review of successes, challenges, and opportunities.
\newblock {\em Diabetes Technology \& Therapeutics}, 18(S2):S2--3, 2016.

\bibitem{rubanova2019latent}
Yulia Rubanova, Ricky~TQ Chen, and David~K Duvenaud.
\newblock Latent ordinary differential equations for irregularly-sampled time series.
\newblock {\em Advances in neural information processing systems}, 32, 2019.

\bibitem{schirmer2022modeling}
Mona Schirmer, Mazin Eltayeb, Stefan Lessmann, and Maja Rudolph.
\newblock Modeling irregular time series with continuous recurrent units.
\newblock In {\em International Conference on Machine Learning}, pages 19388--19405. PMLR, 2022.

\bibitem{shang2017vigan}
Chao Shang, Aaron Palmer, Jiangwen Sun, Ko-Shin Chen, Jin Lu, and Jinbo Bi.
\newblock Vigan: Missing view imputation with generative adversarial networks.
\newblock In {\em 2017 IEEE International conference on big data (Big Data)}, pages 766--775. IEEE, 2017.

\bibitem{shen2023non}
Lifeng Shen and James Kwok.
\newblock Non-autoregressive conditional diffusion models for time series prediction.
\newblock {\em arXiv preprint arXiv:2306.05043}, 2023.

\bibitem{Si2020DeepRL}
Yuqi Si, Jingcheng Du, Zhao Li, Xiaoqian Jiang, Timothy~A. Miller, Fei Wang, W.~Jim Zheng, and Kirk Roberts.
\newblock Deep representation learning of patient data from electronic health records (ehr): A systematic review.
\newblock {\em Journal of biomedical informatics}, page 103671, 2020.

\bibitem{physionetchallenge}
Ikaro Silva, George Moody, Daniel Scott, Leo Celi, and Roger Mark.
\newblock Predicting in-hospital mortality of icu patients: The physionet/computing in cardiology challenge 2012.
\newblock {\em Computational Cardiology}, 39:245–248, 2012.

\bibitem{song2020score}
Yang Song, Jascha Sohl-Dickstein, Diederik~P Kingma, Abhishek Kumar, Stefano Ermon, and Ben Poole.
\newblock Score-based generative modeling through stochastic differential equations.
\newblock In {\em International Conference on Learning Representations}, 2020.

\bibitem{stankeviciute2021conformal}
Kamile Stankeviciute, Ahmed M~Alaa, and Mihaela van~der Schaar.
\newblock Conformal time-series forecasting.
\newblock {\em Advances in neural information processing systems}, 34:6216--6228, 2021.

\bibitem{suo2020glima}
Qiuling Suo, Weida Zhong, Guangxu Xun, Jianhui Sun, Changyou Chen, and Aidong Zhang.
\newblock Glima: Global and local time series imputation with multi-directional attention learning.
\newblock In {\em 2020 IEEE International Conference on Big Data (Big Data)}, pages 798--807. IEEE, 2020.

\bibitem{tang2020rethinking}
Wensi Tang, Guodong Long, Lu~Liu, Tianyi Zhou, Jing Jiang, and Michael Blumenstein.
\newblock Rethinking 1d-cnn for time series classification: A stronger baseline.
\newblock {\em arXiv preprint arXiv:2002.10061}, pages 1--7, 2020.

\bibitem{tashiro2021csdi}
Yusuke Tashiro, Jiaming Song, Yang Song, and Stefano Ermon.
\newblock {CSDI}: Conditional score-based diffusion models for probabilistic time series imputation.
\newblock {\em Advances in Neural Information Processing Systems}, 34:24804--24816, 2021.

\bibitem{heartattack}
Kristian Thygesen, Joseph~S Alpert, Allan~S Jaffe, Bernard~R Chaitman, Jeroen~J Bax, David~A Morrow, and Harvey~D White.
\newblock Fourth universal definition of myocardial infarction (2018).
\newblock {\em European Heart Journal}, 40(3):237--269, 2019.

\bibitem{van2023yet}
Robin van~de Water, Hendrik Schmidt, Paul Elbers, Patrick Thoral, Bert Arnrich, and Patrick Rockenschaub.
\newblock Yet another icu benchmark: A flexible multi-center framework for clinical ml.
\newblock {\em arXiv preprint arXiv:2306.05109}, 2023.

\bibitem{mortality}
William P. T.~M. van Doorn, Patricia~M. Stassen, Hella~F. Borggreve, Maaike~J. Schalkwijk, Judith Stoffers, Otto Bekers, and Steven J.~R. Meex.
\newblock A comparison of machine learning models versus clinical evaluation for mortality prediction in patients with sepsis.
\newblock {\em PLoS One}, 16(1):e0245157, 2021.

\bibitem{van1976stochastic}
Nicolaas~G Van~Kampen.
\newblock Stochastic differential equations.
\newblock {\em Physics reports}, 24(3):171--228, 1976.

\bibitem{vaswani2017attention}
Ashish Vaswani, Noam Shazeer, Niki Parmar, Jakob Uszkoreit, Llion Jones, Aidan~N Gomez, {\L}ukasz Kaiser, and Illia Polosukhin.
\newblock Attention is all you need.
\newblock {\em Advances in neural information processing systems}, 30, 2017.

\bibitem{wahl2018artificial}
Brian Wahl, Aline Cossy-Gantner, Stefan Germann, and Nina~R Schwalbe.
\newblock Artificial intelligence (ai) and global health: how can ai contribute to health in resource-poor settings?
\newblock {\em BMJ global health}, 3(4), 2018.

\bibitem{wang2024deep}
Jun Wang, Wenjie Du, Wei Cao, Keli Zhang, Wenjia Wang, Yuxuan Liang, and Qingsong Wen.
\newblock Deep learning for multivariate time series imputation: A survey.
\newblock {\em arXiv preprint arXiv:2402.04059}, 2024.

\bibitem{wang2015imaging}
Zhiguang Wang and Tim Oates.
\newblock Imaging time-series to improve classification and imputation.
\newblock In {\em Proceedings of the 24th International Conference on Artificial Intelligence}, pages 3939--3945, 2015.

\bibitem{wu2023timesnet}
Haixu Wu, Tengge Hu, Yong Liu, Hang Zhou, Jianmin Wang, and Mingsheng Long.
\newblock Timesnet: Temporal 2d-variation modeling for general time series analysis.
\newblock In {\em The Eleventh International Conference on Learning Representations}, 2023.

\bibitem{wu2020connecting}
Zonghan Wu, Shirui Pan, Guodong Long, Jing Jiang, Xiaojun Chang, and Chengqi Zhang.
\newblock Connecting the dots: Multivariate time series forecasting with graph neural networks.
\newblock In {\em Proceedings of the 26th ACM SIGKDD international conference on knowledge discovery \& data mining}, pages 753--763, 2020.

\bibitem{xu2023density}
Jingwen Xu, Fei Lyu, and Pong~C Yuen.
\newblock Density-aware temporal attentive step-wise diffusion model for medical time series imputation.
\newblock In {\em Proceedings of the 32nd ACM International Conference on Information and Knowledge Management}, pages 2836--2845, 2023.

\bibitem{yan2022neuro}
Ruixuan Yan, Tengfei Ma, Achille Fokoue, Maria Chang, and Agung Julius.
\newblock Neuro-symbolic models for interpretable time series classification using temporal logic description.
\newblock In {\em 2022 IEEE International Conference on Data Mining (ICDM)}, pages 618--627. IEEE, 2022.

\bibitem{Yang2022DeepMPMAM}
Fan Yang, Jian Zhang, Wanyi Chen, Yongxuan Lai, Ying Wang, and Quan Zou.
\newblock Deepmpm: a mortality risk prediction model using longitudinal ehr data.
\newblock {\em BMC Bioinformatics}, 23, 2022.

\bibitem{maskingtraffic2}
Yongchao Ye, Shiyao Zhang, and James J.~Q. Yu.
\newblock Spatial-temporal traffic data imputation via graph attention convolutional network.
\newblock In Igor Farka{\v{s}}, Paolo Masulli, Sebastian Otte, and Stefan Wermter, editors, {\em Artificial Neural Networks and Machine Learning -- ICANN 2021}, pages 241--252, 2021.

\bibitem{yildiz2022multivariate}
A~Yark{\i}n Y{\i}ld{\i}z, Emirhan Ko{\c{c}}, and Aykut Ko{\c{c}}.
\newblock Multivariate time series imputation with transformers.
\newblock {\em IEEE Signal Processing Letters}, 29:2517--2521, 2022.

\bibitem{yoon2017multi}
Jinsung Yoon, William~R Zame, and Mihaela van~der Schaar.
\newblock Multi-directional recurrent neural networks: A novel method for estimating missing data.
\newblock In {\em Time series workshop in international conference on machine learning}, 2017.

\bibitem{zeng2023transformers}
Ailing Zeng, Muxi Chen, Lei Zhang, and Qiang Xu.
\newblock Are transformers effective for time series forecasting?
\newblock In {\em Proceedings of the AAAI conference on artificial intelligence}, volume~37, pages 11121--11128, 2023.

\bibitem{zhang2023comprehensive}
Yongshun Zhang, Xinhang Li, Zehong Li, Wei Liu, Zhihua Zhang, and Jing Liu.
\newblock A comprehensive survey on deep learning for missing data imputation: Taxonomy, challenges, and future directions.
\newblock {\em Information Fusion}, 93:101796, 2023.

\bibitem{zhao2023deep}
Min Zhao, Jianqing Yang, Wei Wang, Xiaojuan Ma, and Zaiqing Nie.
\newblock Deep learning using 2d convolutional neural networks for multivariate clinical time series: A review and comparison.
\newblock {\em IEEE Journal of Biomedical and Health Informatics}, 27(4):1723--1734, 2023.

\bibitem{zhou2020graph}
Jie Zhou, Ganqu Cui, Shengding Hu, Zhengyan Zhang, Cheng Yang, Zhiyuan Liu, Lifeng Wang, Changcheng Li, and Maosong Sun.
\newblock Graph neural networks: A review of methods and applications.
\newblock {\em AI open}, 1:57--81, 2020.

\bibitem{correlation3}
Y.~Zhou, J.~Shi, R.~Stein, X.~Liu, R.~N. Baldassano, C.~B. Forrest, Y.~Chen, and J.~Huang.
\newblock Missing data matter: an empirical evaluation of the impacts of missing ehr data in comparative effectiveness research.
\newblock {\em J Am Med Inform Assoc}, 30(7):1246--1256, Jun 2023.

\bibitem{zhuang2023mixture}
Ningning Zhuang, Mengmeng Gong, Jiahui Zhu, Yilong Yin, Bo~Liu, and Jun Wang.
\newblock On mixture density networks: A survey.
\newblock {\em IEEE Transactions on Neural Networks and Learning Systems}, 2023.

\end{thebibliography}



\end{document}